\documentclass{ieeeaccess}
\usepackage{cite}
\usepackage{amsmath,amssymb,amsfonts}
\usepackage{algorithmic}
\usepackage{graphicx}
\usepackage{textcomp}
\usepackage{caption}

\UseRawInputEncoding

\usepackage{times}
\usepackage{epsfig}
\usepackage{url}

\usepackage{ctable} 

\def\BibTeX{{\rm B\kern-.05em{\sc i\kern-.025em b}\kern-.08em
    T\kern-.1667em\lower.7ex\hbox{E}\kern-.125emX}}
\begin{document}
\history{Date of publication xxxx 00, 0000, date of current version xxxx 00, 0000.}
\doi{10.1109/ACCESS.2017.DOI}

\title{A Survey of Super-Resolution in Iris Biometrics with Evaluation of Dictionary-Learning
}

\author{\uppercase{Fernando Alonso-Fernandez}\authorrefmark{1},
\IEEEmembership{Member, IEEE},\uppercase{Reuben A. Farrugia}\authorrefmark{2},\IEEEmembership{Senior Member, IEEE},
\uppercase{Josef Bigun}\authorrefmark{1},\IEEEmembership{Fellow, IEEE},
\uppercase{Julian Fierrez}\authorrefmark{3},\IEEEmembership{Member, IEEE},
and \uppercase{Ester Gonzalez-Sosa}\authorrefmark{4}}
\address[1]{School of Information Technology (ITE), Halmstad University, Box 823, Halmstad SE 301-18, Sweden (e-mail: feralo@hh.se, josef.bigun@hh.se)}
\address[2]{Department of Communications and Computer Engineering (CCE), University of Malta, Malta (e-mail: reuben.farrugia@um.edu.mt)}
\address[3]{Biometrics and Data Pattern Analytics Lab (BiDA) - ATVS, Escuela Politecnica Superior, Universidad Autonoma de Madrid, 28049 Madrid, Spain (e-mail: julian.fierrez@uam.es)}
\address[4]{Nokia Bell-Labs, 28045 Madrid, Spain (e-mail: ester.gonzalez@nokia-bell-labs.com)}

\tfootnote{This work was initiated while F. A.-F. was a visiting researcher at the University
of Malta, funded by EU COST Action IC1106.
F. A.-F and J. B. authors thank the Swedish Research Council, the Swedish Innovation Agency, and
the CAISR / SIDUS-AIR projects of the Swedish Knowledge Foundation.
J. F. author is funded by project CogniMetrics (TEC2015-70627-R) from Spanish MINECO/FEDER.}

\markboth
{Author \headeretal: Preparation of Papers for IEEE TRANSACTIONS and JOURNALS}
{Author \headeretal: Preparation of Papers for IEEE TRANSACTIONS and JOURNALS}

\corresp{Corresponding author: Fernando Alonso-Fernandez (e-mail: feralo@hh.se).}

\begin{abstract}

The lack of resolution has a negative impact on the performance of image-based biometrics.
%
%
%
%
While many generic super-resolution methods have been proposed to restore low-resolution images,
they usually aim to enhance their visual appearance.
However, an overall visual enhancement of biometric images does not necessarily correlate with a better recognition performance.
Reconstruction approaches need thus to incorporate specific information from the target biometric modality to effectively improve recognition performance.
This paper presents a comprehensive survey of iris super-resolution approaches proposed in the literature.
%
We have also adapted an Eigen-patches reconstruction method based on PCA Eigen-transformation of local image patches.
%
The structure of the iris is exploited by building a patch-position dependent dictionary.
In addition, image patches are restored separately, having their own reconstruction weights. This allows the solution to be locally optimized, helping to preserve local information.
%
%
%
To evaluate the algorithm, we degraded high-resolution images from the CASIA Interval V3 database.
%
Different restorations were considered, with $15\times15$ pixels being the smallest resolution evaluated.
To the best of our knowledge, this is among the smallest resolutions employed in the literature.
%
The experimental framework is complemented with six publicly available iris comparators,
which were used to carry out biometric verification and identification experiments.
Experimental results show that the proposed method 
significantly outperforms both bilinear and bicubic interpolation at very low-resolution. The performance of a number of comparators attain an impressive Equal Error Rate as low as 5\%, and a Top-1 accuracy of 77-84\% when considering iris images of only $15 \times 15$ pixels.
These results clearly demonstrate the benefit of using trained super-resolution techniques to improve the quality of iris images prior to matching.

\end{abstract}


\titlepgskip=-15pt

\maketitle

\section{Introduction}
%
%

Iris recognition systems are known to achieve very high accuracy when captured in controlled environments and using the near infrared (NIR) spectrum.
Nevertheless, recognition in applications such as
mobile biometrics, surveillance and recognition at a distance
has not reached the same level of maturity \cite{[Nigam15]}.
In these environments, the acquisition cannot be controlled,
and performance can significantly drop due to the lack of pixel resolution \cite{[Jain11a]}.
%
%
Furthermore, smart cards or remote applications may further reduce the quality of the image using JPEG2000 compression \cite{[Quinn14]}.
%
%
In this context, super-resolution techniques can be used to enhance the quality of low resolution images, in order to improve the recognition performance of biometric systems \cite{[Nguyen18a]}.

\begin{figure*}[htb]
     \centering
     \includegraphics[width=.95\textwidth]{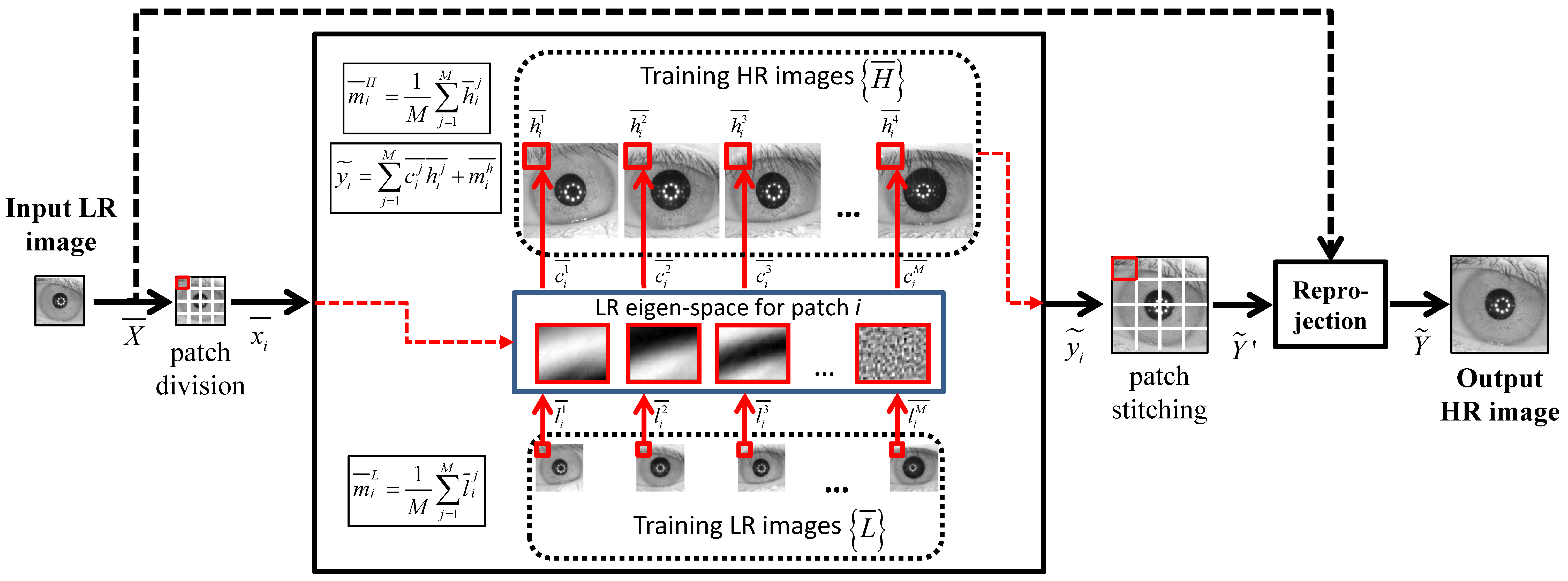}
     \caption{Structure of the eigen-patch hallucination system.}
     \label{fig:system}
\end{figure*}

Two main categories of super-resolution methods are usually distinguished in the literature \cite{[Park03]}:
reconstruction-based and learning-based methods.
Reconstruction-based methods register and fuse a number of consecutive
low-resolution images to estimate the high-resolution image.
These methods are known to achieve relatively small magnification factors and are most suitable to restore static and non-rigid objects.
On the other hand, learning-based methods use coupled training dictionaries
to learn the mapping relations between low- and high-resolution
image pairs. 
%
Learning-based methods have the advantage of estimating the high-resolution image using only one low-resolution image as input, and they are also known to achieve higher magnification factors\cite{[Park03]}.

In recent years, there has been an increased interest in the application of super-resolution
to different biometric modalities, such as face, iris, gait or fingerprint \cite{[Nguyen18a]}.
However, despite the vast literature of super-resolution methods 
\cite{[Nasrollahi2014],[Yue16]},
such techniques are designed to restore generic images. They do not exploit the specific structure of biometric images, which causes the solution to be sub-optimal \cite{[Farrugia17]}.
Instead, they try to improve visual clarity and perception, usually by optimizing image fidelity measures such as the Peak Signal-to-Noise Ratio (PSNR) or the Structural Similarity (SSIM) index. But improving the visual quality of biometric images does not necessarily correlate with a better recognition performance \cite{[Alonso12a],[Nguyen18a]}, which is the ultimate aim of applying super-resolution to biometrics \cite{[Nguyen18a]}.
Thus, adaptation
of super-resolution techniques to the particularities of images
from a specific biometric modality is needed to achieve
a more efficient upsampling \cite{[Baker02]}.

\begin{figure*}[htb]
     \centering
     \includegraphics[width=.7\textwidth]{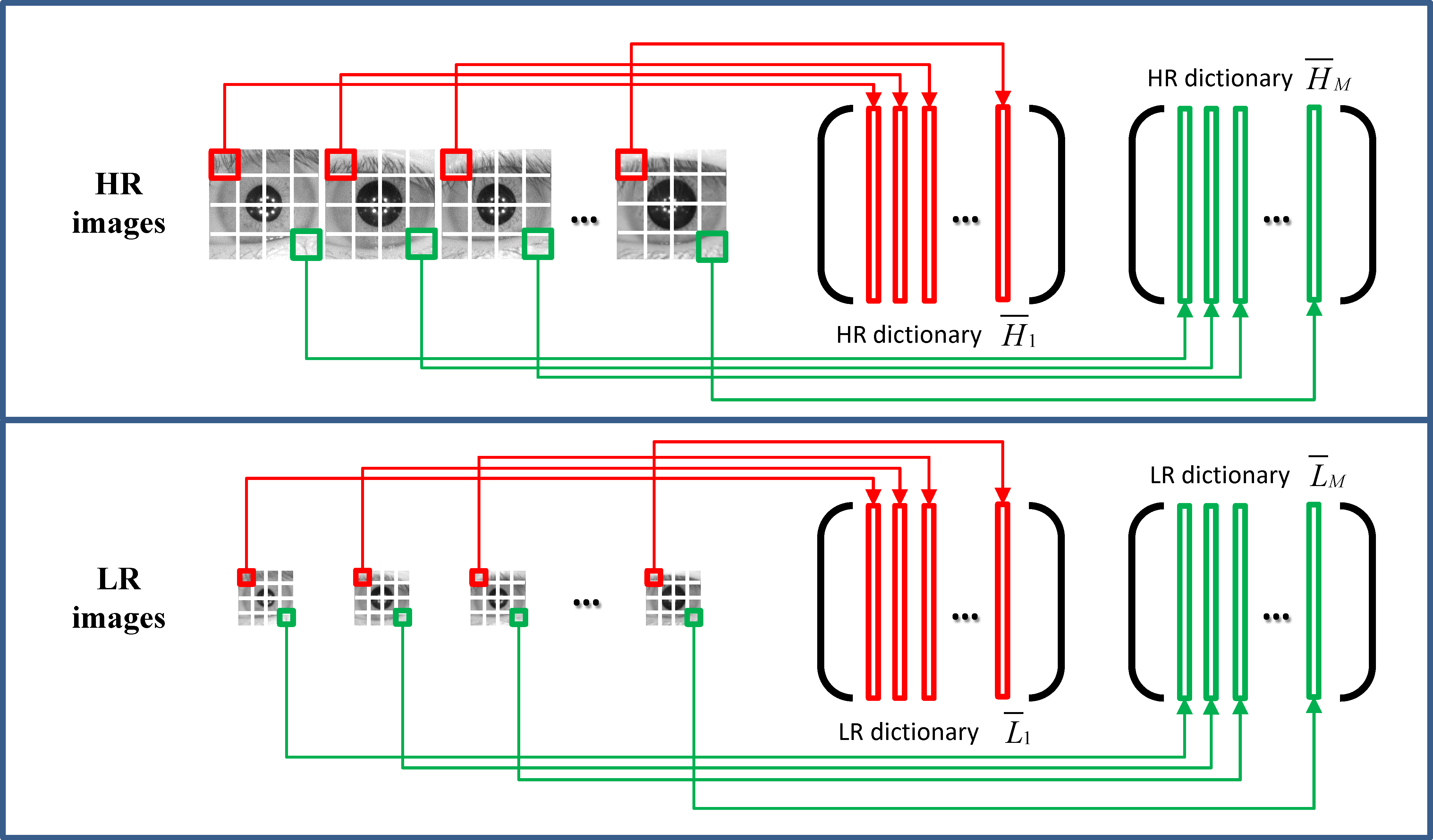}
     \caption{Dictionary construction of coupled low- and high-resolution patches using the position-patch principle.}
     \label{fig:dictionary}
\end{figure*}

Consequently, this paper addresses
the problem of restoring the resolution by exploiting the structure of the iris to improve recognition performance.
After a comprehensive survey of the literature in super-resolution applied to iris biometrics,
we investigate
the use of local iris super-resolution
based on Principal Component Analysis (PCA).
In this learning-based approach,
an Eigen-transformation is computed on each local patch of the input low-resolution image (Figure~\ref{fig:system}).
For this purpose, a dictionary database of coupled low- and high-resolution patches is employed (Figure~\ref{fig:dictionary}).
Given a low-resolution patch, it is projected onto
a low-dimensional subspace which captures most of the information contained in the patch.
The low-dimensional eigen-space for each patch position is computed by applying PCA to the set of collocated low-resolution patches of the training dictionary.
The high-resolution patch is then reconstructed by linear combination of
the collocated high-resolution patches of the dictionary.
It is important to emphasize that each patch, which caters for a specific region of the iris, has its own distinct coupled dictionaries.
Also, each input patch is allowed to have its own reconstruction weights,
so the solution is locally optimized.
Reconstructing each patch separately, with its own optimum weights,
allows to better recover local texture details.
This is essential due to the prevalence of texture-based methods in ocular biometrics \cite{[Nigam15]}.

The present paper extends previous studies \cite{[Alonso15b],[Alonso16b]}
with additional experiments.
A related method was proposed and studied for face super-resolution
by Chen and Chien \cite{[Chen14]},
which was the initial source that motivated the method studied here.
The proposed iris super-resolution method was evaluated using a dataset of 1,872 iris images captured using a near infrared sensor.
High-resolution images, with a size of 231$\times$231 pixels and an average iris diameter of 210 pixels, were down-sampled to different scales, with the smallest resolution being of 15$\times$15 pixels.
Such simulated downsampling is a common approach in the literature, due to the lack of databases containing very low-resolution images and their corresponding high-resolution reference images \cite{[Wang14]}.
In addition to traditional image fidelity metrics between reconstructed and
reference high-resolution images,
in this paper we also report biometric verification and identification experiments using reconstructed iris images.
To the best of our knowledge, this is one of the few iris super-resolution studies which reports identification experiments.
In comparison to \cite{[Alonso15b],[Alonso16b]},
we also incorporate four new iris comparators to our experimental framework
\cite{[Rathgeb10],[Monro07],[Ko07],[Ma03]}
in addition to the two previously employed \cite{[MasekThesis03],[Lowe04]}.
%
%

Simulation experiments conducted in this paper show that the proposed method achieves accuracies superior to those attained using bilinear and bicubic interpolation schemes. 
It is observed that recognition rates degrade more
rapidly with both bilinear and bicubic interpolation as resolution
decreases.
At the smallest iris resolution (15$\times$15 pixels), two particular comparators stand out,
with EER values of $\sim$5\% and a Top-1 accuracy of 77-84\% in this extreme case.
It is also shown that recognition performance
is not significantly degraded with any given comparator until a resolution of only 29$\times$29 pixels when using our proposed PCA iris super-resolution method.
This allows to reduce the storage or data transmission requirements, or to increase the distance between the user and the iris sensor, two important requirements for biometric technologies
to achieve massive adoption \cite{[Jain11a]}.
We have also observed that, despite iris images reconstructed with PCA have better subjective quality, the image fidelity measures employed (PSNR and SSIM) do not have the same sensitivity to reductions in resolution. This is also acknowledged in other studies which have pointed out that image fidelity metrics do not behave equally under the same image degradation \cite{[AlNajjar12],[Hore10]}. 
In addition, each comparator has different behaviour when resolution is reduced. Most of the comparators show a stable authentication performance until a certain resolution is employed, but the cut-off resolution is different for each one. On the other hand, one particular comparator does not suffer a significant degradation in performance, giving a consistent accuracy across nearly all resolutions.
These differences in behaviour of image fidelity metrics and biometric comparators among themselves highlight the necessity of adapting super-resolution techniques to cater for the
particularities of a specific biometric modality \cite{[Baker02]}.
It is also crucial not to assess only the fidelity of the reconstruction in the visual sense,
but to evaluate the capability of the reconstruction algorithm to improve authentication performance with the particular recognition features to be employed
\cite{[Grother07],[Alonso12a]}.
%

\begin{table*}
\scriptsize
\begin{center}
\begin{tabular}[htb]{|c|c|c|c|c|c|c|c|c|c|c|}


\multicolumn{11}{c}{\textbf{PIXEL DOMAIN: RECONSTRUCTION-BASED}} \\   \specialrule{.2em}{.1em}{.1em}  \hline

\textbf{} & \textbf{} & \textbf{Input} & \textbf{Patch} & \textbf{} & \textbf{Input} &\textbf{Smallest} &\textbf{Simulated} & \textbf{Recognition} & \multicolumn{2}{c|}{\textbf{Accuracy}} \\ \cline{10-11}

\textbf{Ref.} & \textbf{Algorithm} & \textbf{Data} & \textbf{Based} & \textbf{Database} & \textbf{Images} & \textbf{LR Size} & \textbf{Downs.} & \textbf{Features} &  EER & Rank-1 \\ \hline \hline

\cite{[Barnard06]}  & Inverse Optimization  &  Iris  & No  & Own videos   &  15   & 30px ED  & no  &  Iris Code  &  n/a & n/a  \\ \hline \hline

\cite{[Fahmy07]}  & Auto-regressive models  &  Iris  & No  & Own videos   &   9  & n/a  & no  &  Iris Code  & n/a & n/a  \\ \hline \hline


\cite{[Nguyen10]}\cite{[Nguyen11]}  & Weighted average  & Polar  & No  & MBGC Portal   & 5  & 90px ED  & no  & Iris Code  & 0.7\% & n/a  \\ \hline \hline

\cite{[Nguyen10a]}  & Robust-mean average  & Polar  & No   & MBGC Portal  &  var  & 90px ED & no  & Iris Code   & 4.1\% & n/a  \\ \hline \hline

\cite{[Hollingsworth09a]}  & Mean/median  & Polar   & No   & MBGC Iris    &   10  & 220px ED &  no &  Iris Code  & 0.7\% & n/a \\ \hline \hline


\cite{[Jillela11]}  & PCT enhance + average  & Polar   & No   & MBGC Iris    &  6   & 20px ED &  yes &  Log-Gabor  & 1.76\% & n/a \\ \hline \hline

\cite{[Ren12]}  & Iterated back projection  & Polar   & No   & CASIA 3.0   &  4   & 128$\times$16 PI & yes  & Iris Code  & 8.69\% & n/a \\ \hline \hline

\cite{[Othman13]}\cite{[Othman15]} & Weighted average  & Polar   & No   & MBGC Portal  &  n/a  & 90px ED &  no & Iris Code   & 2.58\% & n/a  \\ \cline{5-10}
                        &   &    &    & Q-FIRE   &  6  & 110px ED &   no & Iris Code   & n/a & n/a \\ \hline \hline


\cite{[Hsieh16]}  & Wavefront coding +   & Polar   & Yes   & Own videos   &  11   & n/a  & no  &  Log-Gabor  & 0\% & n/a  \\

& exp-weighted average  &    &  & @3m &   &   &   &   &  &   \\ \hline \hline


\cite{[Deshpande17a]} & Gaussian Process   & Polar   & Yes   &  CASIA   &  n/a   & 300$\times$40 PI & no  & GLCM, moments,   &  n/a & n/a  \\

   & Regression &    &    &   Long Range   &      &  &  & statistical features    & &   \\ \hline \hline

\cite{[Deshpande17]}  & Total Variation  & Polar   & No   &  CASIA     &  6   & 300$\times$40 PI & no  &   GLCM   & n/a & n/a \\

  &  &    &    & Long Range   &    &   &    &    & &  \\  \hline \specialrule{.2em}{.1em}{.1em}

%

\multicolumn{11}{c}{} \\ 

\multicolumn{11}{c}{\textbf{PIXEL DOMAIN: LEARNING-BASED}} \\  \specialrule{.2em}{.1em}{.1em}   \hline

\textbf{} & \textbf{} & \textbf{Input} & \textbf{Patch} & \textbf{} & \textbf{Input} &\textbf{Smallest} &\textbf{Simulated} & \textbf{Recognition} & \multicolumn{2}{c|}{\textbf{Accuracy}} \\ \cline{10-11}

\textbf{Ref.} & \textbf{Algorithm} & \textbf{Data} & \textbf{Based} & \textbf{Database} & \textbf{Images} & \textbf{LR Size} & \textbf{Downs.} & \textbf{Features} &  EER & Rank-1 \\ \hline \hline

\cite{[Huang03]}  & High frequency inference  &  Polar  & Yes   & CASIA    &  1   & n/a  & yes &  Circular Filters  & 15\% & 89.7\% \\ \hline  \hline

\cite{[Shin09]}  & Multi-layer Perceptrons  &  Iris  &  Yes  & CASIA Interval     & 1   & 53px ED & yes &  Iris Code   & 1.39\% & n/a  \\ \hline \hline

\cite{[Liu14]}  & Score-level mapping  & Polar   & Yes   & Q-FIRE    &  1   & 110px ED   & no & Ordinal Measures  & 1.6\% & n/a \\ \hline \hline

\cite{[Aljadaany15]}  & Bayesian Modelling +    &  Polar  & Yes   & CASIA   &   1   & n/a  & yes  & n/a   & 2\% & n/a \\

 & Sparse Representation   &    &   &   &   &     &     &  & & \\ \hline \hline

\cite{[Alonso15b]}\cite{[Alonso16b]}
  & PCA modelling  & Iris   &  Yes  & CASIA Interval   &    1  & 11px ED & yes  &  Log-Gabor, SIFT   & 6.44\% & n/a  \\ \cline{1-1}  \cline{7-8}  \cline{9-11}

This paper &   &    &    &    &   & 13px ED   & yes &  6 comparators   & 4.79\% & 84.2\%  \\ \hline \hline

\cite{[Alonso17]}  & Neighbour Embedding  &  Iris  & Yes   & CASIA Interval    &  1   & 13px ED  & yes   & Log-Gabor, SIFT   & 3.58\% & n/a \\ \hline \hline

\cite{[Alonso17a]}
  & PCA modelling  & Iris   &  Yes  & VSSIRIS    &  1   & 13px ED &  yes &  Log-Gabor, SIFT  & 4.1\% & n/a  \\

  & Neighbour Embedding &    &    &   &   &    &   &    &  &  \\ \hline \hline

\cite{[Zhang16a]}  & Convolutional Networks   &  Polar  & Yes   & CASIA Mob 1    &   1  & 110px ED    & no & Ordinal Measures   &  3.61\% & n/a \\ \cline{5-8} \cline{10-11}

 & Random Forests   &    &   & CASIA Mob 2  & 1  &   132px ED & no  &   & 1.82\% & n/a \\ \hline \hline

\cite{[Ribeiro17]}  & Convolutional Networks  & Iris   & Yes   & CASIA Interval    &  1   & 13px ED  & yes &  Log-Gabor, SIFT   &   6.26\% & n/a \\

  & Stacked Auto-Encoders  &    &   &     &   &     & &     & &  \\  \hline \hline

\cite{[Ribeiro18]}  & Convolutional Networks  & Iris   & Yes   & CASIA Interval    &  1   & 13px ED  & yes  & CG, QSW, SIFT   &   27.6\% & n/a  \\ \cline{5-8} \cline{10-11}

  &   &    &   &  VSSIRIS & 1  &  13px ED & yes  &  &   12\% & n/a    \\  \hline \specialrule{.2em}{.1em}{.1em}

\multicolumn{11}{c}{} \\ 

\multicolumn{11}{c}{\textbf{FEATURE DOMAIN: LEARNING-BASED}} \\ 
\specialrule{.2em}{.1em}{.1em}  \hline


\textbf{} & \textbf{} & \textbf{Input} & \textbf{Patch} & \textbf{} & \textbf{Input} &\textbf{Smallest} &\textbf{Simulated} & \textbf{Recognition} & \multicolumn{2}{c|}{\textbf{Accuracy}} \\ \cline{10-11}

\textbf{Ref.} & \textbf{Algorithm} & \textbf{Data} & \textbf{Based} & \textbf{Database} & \textbf{Images} & \textbf{LR Size} & \textbf{Downs.} & \textbf{Features} &  EER & Rank-1 \\ \hline \hline

\cite{[Nguyen11a]}  & Bayes MAP  & Eigen-Iris   & No   & MBGC Iris    &   1   & 50px ED & yes &  Eigen-Iris   & 4.5\% & n/a \\ \hline \hline

\cite{[Nguyen12]}\cite{[Nguyen13]}  & Bayes MAP  & Gabor   & No   & MBGC Portal    &   5   & 90px ED & no  & Iris Code    & 0.5\% & n/a \\ \hline \hline

\cite{[Liu13]}  & Markov Networks  & Iris Code   & No   &  Q-FIRE    &  n/a   & 110px ED & no  & Ordinal Measures   & 2.6\% & n/a \\  \hline \specialrule{.2em}{.1em}{.1em}



\multicolumn{8}{c}{} \\

\end{tabular}
\caption{Overview of existing iris super-resolution works. Smallest low-resolution (LR) size refers to the smallest size of the input data used in the
reported experiments (ED=Eye Diameter, refers to the average diameter of the iris in pixels; 
PI=Polar Image).
Simulated downsampling indicates if the images employed in the study have been down-sampled from high-resolution reference images.
%
%
The reported accuracy corresponds to the best accuracy obtained with the smallest iris size (shown in column 7).
In all cases, near-infrared (NIR) data is used, except in
the works \cite{[Fahmy07],[Hsieh16],[Alonso17a],[Ribeiro18]}. All other terms are explained in the text or in referenced papers.} \label{tabla:SOA}
\end{center}
\end{table*}
\normalsize

\begin{figure*}[htb]
     \centering
     \includegraphics[width=.9\textwidth]{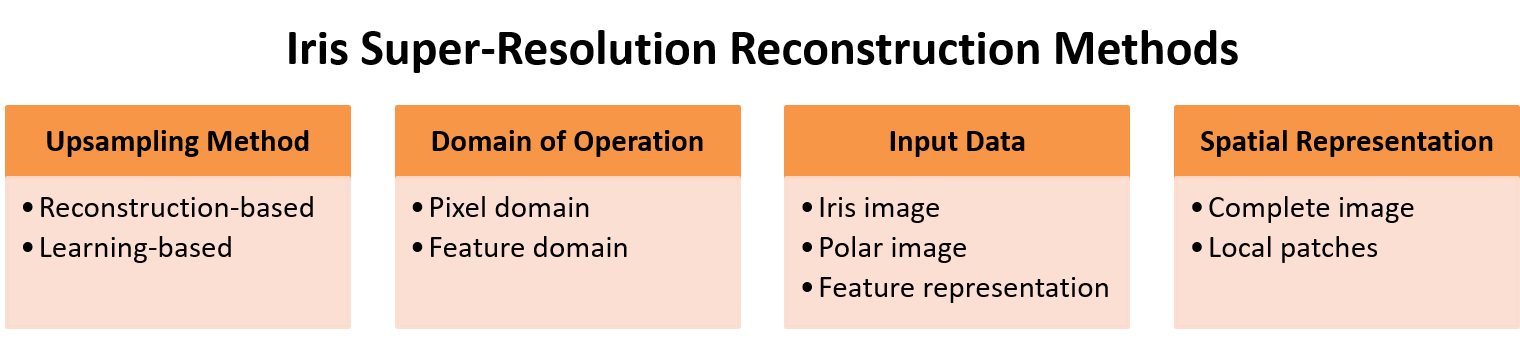}
     \caption{Taxonomy of existing iris super-resolution reconstruction methods.}
     \label{fig:taxonomy}
\end{figure*}

The rest of the paper is organized as follows.
The remaining of this section summarizes 
the main contributions of this paper.
%
%
%
%
In Section~\ref{sect:soa}, we provide a comprehensive overview of the application
of super-resolution techniques to iris biometrics.
This is followed in Section~\ref{sect:method} by the description of
the proposed super-resolution algorithm, which adopts low- and high-resolution coupled dictionaries to learn an optimal up-scaling function for each patch.
Then, this super-resolution algorithm is studied for iris biometrics.
While this method is employed for iris super-resolution, it is general enough to be applied to other biometric modalities.
%
The experimental framework, including database,
protocol, and iris recognition algorithms employed,
is given in Section~\ref{sect:framework},
while results are given in Section~\ref{sect:results}.
Conclusions are finally drawn in Section~\ref{sect:conclusions}.

\subsection{Contributions}

The contributions of this paper are as follows:

\begin{itemize}
  \item \textit{A survey of super-resolution applied to iris biometrics}
  (Table~\ref{tabla:SOA}).
We give a comprehensive overview of references found in the literature (we focus primarily
on papers that appeared in IEEE Xplore, ScienceDirect or SpringerLink,
as these appear to currently be the major sources of
publications in the biometrics field).
We provide a basic algorithmic descriptions of each approach, including
the database employed for its evaluation, the smallest size of input low-resolution images considered,
the features used for recognition experiments, and the reported recognition results (if any).
We 
also provide a taxonomy of existing iris super-resolution methods
based on different factors (Figure~\ref{fig:taxonomy}),
which include
the domain of operation (pixel or feature domain),
the input data source (iris image, polar image, or feature representations),
or the spatial representation (if the method uses complete images or local image patches to carry out the reconstruction).
%
\end{itemize}

\begin{itemize}
  \item \textit{A generic super-resolution method} which employs low- and high-resolution coupled dictionaries to learn the optimal up-scaling function for each patch. This approach is able to recover important texture detail which is essential for most biometric recognition systems, including iris.
Following the work of Chen and Chien initially developed for face biometrics \cite{[Chen14]},
a PCA Eigen-transformation is computed on local patches of the input low-resolution image.
The high-resolution patch is then hallucinated as a linear combination of collocated high-
resolution patches contained in the training dictionary.
This way, every patch has its own optimal coefficients, which allows to reconstruct patches that are locally optimal and thus, to recover more texture detail.
%
We evaluate our general super-resolution approach using iris images of resolution as low as 15$\times$15 pixels
(corresponding to an average iris diameter $\sim$13 pixels). To the best of our knowledge, this iris size
is much smaller than any other work reported in the literature, apart from ours (see Table~\ref{tabla:SOA}).
%
Another benefit is that, unlike other methods that restore the normalized polar image, our method is agnostic of the feature extraction method used, since it is applied directly on the iris low-resolution image. This makes the proposed method generic and independent from the iris comparator used. It also allows the use of features which are extracted directly from iris images
without conversion to polar coordinates, e.g. \cite{[Alonso09],[Nguyen18]}.
\end{itemize}

\begin{itemize}
  \item  \textit{Multi-algorithmic evaluation}. In our previous works \cite{[Alonso15b],[Alonso16b]},
we used only two iris comparators for the experimental study. Here, we use six different publicly available iris feature
extraction methods from popular and state-of-the-art schemes \cite{[Rathgeb13]} based on
1D log-Gabor filters \cite{[MasekThesis03]},
the SIFT operator \cite{[Lowe04],[Alonso09]},
%
%
local intensity variations in iris textures \cite{[Rathgeb10]},
the Discrete-Cosine Transform \cite{[Monro07]},
cumulative-sum-based grey change analysis \cite{[Ko07]}, and
Gabor spatial filters \cite{[Ma03]}.
The SIFT method exploits local features
where discrete key-points are extracted directly from the iris region,
while the other methods exploit other texture properties from the
iris polar image computed according to Daugman's rubber sheet model
\cite{[Daugman04]}.
%
%
\end{itemize}

\begin{itemize}
  \item  \textit{Comprehensive evaluation on a database of near-infrared iris images}.
We employ in our experiments 1,872 images from the CASIA-Iris Interval v3 database of
the Institute of Automation, Chinese Academy of Sciences (CASIA) \cite{[CASIAdb]}.
High-resolution images, of size 231$\times$231 pixels and an
average iris diameter of 210 pixels,
are sub-sampled to reduce their size by 1/2, 1/4, 1/6, 1/8, 1/10, 1/12, 1/14 and 1/16. The latter corresponds to an image
size of just 15$\times$15 pixels and an average iris diameter of $\sim$13 pixels.
The performance of the iris super-resolution algorithm is measured in terms of PSNR and
SSIM full reference metrics, which compute the fidelity between the original high-resolution image and the restored ones.
Moreover, we carry out verification and identification experiments with the mentioned iris recognition algorithms, being one of the few studies in the literature that reports identification experiments.
This is also the most extensive and up-to-date experimental framework in the context of
iris super-resolution literature, providing extensive validation experiments.
\end{itemize}

\section{Super-Resolution for Iris Biometrics}
\label{sect:soa}

Super-resolution (SR) techniques aim to recover the missing high
resolution (HR) image $\overline Y$ given a low-resolution (LR)
image $\overline X$.
The low-resolution image 
is considered as a warped, blurred and down-sampled version
of the high-resolution image. 
This can be mathematically expressed using

\begin{equation}
\overline{X} = D B W \overline{Y} + \overline n
\label{eq:acquisition-full}
\end{equation}

\noindent where $W$ is the warping matrix,
$B$ is the blurring kernel
(also called point spread function in some studies),
$D$ is the downsampling matrix,
and $\overline n$ represents additive noise.
For simplicity, some works omit the warp matrix and noise, leading to

\begin{equation}
\overline{X} = D B \overline{Y}
\label{eq:acquisition}
\end{equation}

\begin{figure*}[htb]
     \centering
     \includegraphics[width=.9\textwidth]{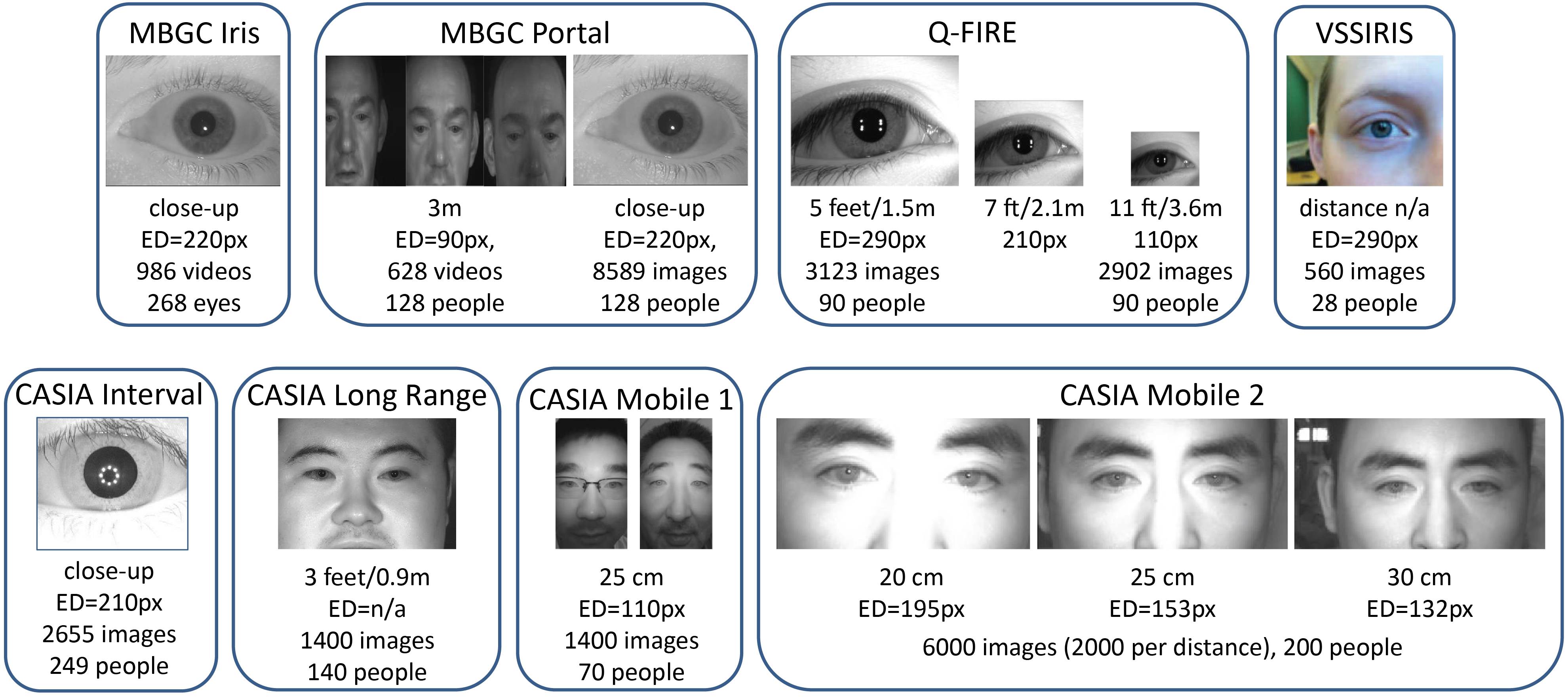}
     \caption{Samples of databases used in iris super-resolution research. Information is also given regarding: the distance between the individual and the acquisition sensor, the number of available images or videos, the number of individuals captured, and the average iris diameter in pixels of the images contained in the database (ED=Eye Diameter). 
     }
     \label{fig:dbs}
\end{figure*}

Super-resolution techniques are classically divided into two categories \cite{[Park03]}: reconstruction- and learning-based methods.
\textit{Reconstruction-based} methods register and fuse a number of
low-resolution images to estimate the high-resolution image.
Several images are aligned and combined in a pixel-wise manner to obtain a reconstructed image.
Given a set of $N$ images $\overline{X}_i$, the super-resolved image
$\tilde{Y}$ is estimated as

\begin{equation}
\tilde{Y}(x,y) = \frac{{\sum\limits_{i = 1}^N {w_i \overline{X}_i(x,y)} }}{{\sum\limits_{i = 1}^N {w_i } }}
\label{eq:reconstruction_fuseim}
\end{equation}

\noindent where $\tilde{Y}(x,y)$ is the intensity value at pixel $(x,y)$ of
the super-resolved image, $\overline{X}_i(x,y)$ is the intensity value at the same location
of the input image $i$, and ${w_i}$ are the combination weights.
While these methods can exploit the correlation and redundancies present in multiple frames to restore the missing detail, they cannot be employed in cases where only one image is available.
Moreover, these methods are known to fail to reconstruct dynamic non-rigid objects and can only achieve small magnification factors.
On the other hand, \textit{learning-based} methods use coupled low- and high-resolution dictionaries
to learn the mapping relations between low- and high-resolution image pairs in order to synthesize a
high-resolution image from the observed low-resolution one.
Learning-based methods have the advantage of needing only one image as input,
and they generally allow to achieve higher magnification factors \cite{[Park03]}.

\subsection{Taxonomy of Iris Super-Resolution Algorithms}

Table~\ref{tabla:SOA} gives a summary in chronological order of existing works on iris super-resolution.
Apart from the distinction between reconstruction- and learning-based methods,
they can be also categorized based on other factors, which are summarized in Figure~\ref{fig:taxonomy}:

\begin{itemize}
  \item 
  Domain of operation (pixel or feature domain).
  The majority of studies work in the \textit{pixel domain} (top and middle part of Table~\ref{tabla:SOA}),
  estimating pixel intensity values of the enhanced image.
  As a result, a new image with improved resolution is produced, which translates
  to a visual related enhancement.
  A few studies carry out the enhancement in the feature space (bottom part of Table~\ref{tabla:SOA}), shifting the reconstruction
  operation from the pixel domain to the \textit{feature domain} employed for recognition
  \cite{[Nguyen11a],[Nguyen12],[Nguyen13],[Liu13]}.
  The latter approaches explicitly aim at improving the recognition performance,
  instead of the visual appearance.
  On the other hand, they have the disadvantage of being tied to a particular feature representation.
  %
\end{itemize}

\begin{itemize}
  \item Type of data used as input for enhancement (iris image, polar image, or feature representation). This is indicated in column 3 of Table~\ref{tabla:SOA}. The majority of pixel-based methods super-resolve the \textit{polar} image directly \cite{[Daugman04]}, while others reconstruct the entire \textit{iris} image. The latter has the advantage of being usable with feature extraction methods that do not employ polar representation \cite{[Alonso09],[Nguyen18]}.
  %
  Feature-based methods, on the other hand, receive as input a \textit{feature} representation of the low-resolution image. Then, instead of producing an enhanced image as output, they estimate a feature representation of the reconstructed image.
\end{itemize}

\begin{itemize}
  \item Spatial representation employed (complete image or local patches).
  This is indicated in column 4 of Table~\ref{tabla:SOA}.
  Some approaches, also called global methods, carry out reconstruction of the \textit{complete image}.
  Patch-based methods, on the other hand, hallucinate \textit{local patches} separately.
  The reconstructed high-resolution patches are then stitched together to form the high-resolution image.
  This allows each patch to have its own optimal reconstruction coefficients,
  providing better quality reconstructed prototypes with better local detail and lower distortion \cite{[Farrugia17]}.
\end{itemize}

In addition to the aforementioned properties, we also provide information in Table~\ref{tabla:SOA} regarding:


\begin{itemize}
  \item The database employed (indicated in column 5). As mentioned in the caption, nearly all studies make use of near-infrared (NIR) data.
  Some sample images from each database are also shown in Figure~\ref{fig:dbs}, together with their most representative information.
  In particular, we indicate: the distance to the acquisition sensor, the number of images or videos available, the number of individuals, and the average iris diameter of the images contained in the database.
  %
  %
  The following is a short description of each database, highlighting its most important features not contained in Figure~\ref{fig:dbs}.
  %
  The Multiple Biometric Grand Challenge Portal database, or \textit{MBGC Portal} \cite{[Phillips09]}, contains
  face video sequences of people walking naturally through a portal located 3m from a fixed-focal-length NIR camera (Pulnix TM-4000CL).
  It also contains iris images of good quality captured with a close-up NIR iris sensor from the same individuals.
  %
  The \textit{MBGC Iris} video database contains videos of irises collected using a close-up NIR iris camera (Iridian LG EOU 2200).
  %
  The \textit{CASIA Long Range} database contains face images captured at 3 feet (0.9 m) with a high resolution NIR camera.
  The \textit{CASIA Iris Interval} database has iris images captured with a close-up NIR sensor.
  %
  The \textit{CASIA Iris Mobile} v1.0 and v2.0 databases contain face images at varios distances captured with a NIR imaging module connected to a smart-phone by USB.
  %
  %
  The \textit{Q-FIRE} database \cite{[Johnson10]} has iris videos captured at 5, 7, and 11 feet away with a Dalsa 4M30 infrared camera and a Tamron AF 70-300mm telephoto lens.
  %
  And finally, the \textit{VSSIRIS} database \cite{[Raja14b]}. This is the only database in visible range, containing images captured using the rear camera of two different smart-phones (Apple iPhone 5S and Nokia Lumia 1020).

\end{itemize}

\begin{itemize}
   \item The number of low-resolution images used by the reconstruction algorithm to generate a high-resolution representation (indicated in column 6). Existing reconstruction-based methods employ a variable number which goes from four to fifteen, while learning-based algorithms in the pixel domain only employ one. It is also found that many learning-based algorithms working in the feature domain employ several images as input. However, since the mapping between low- and high-resolution manifolds is learned, we classify these methods as learning-based. Indeed, nothing prevents learning-based methods to employ more than one image, although one of their main advantages is that they can generate a high-resolution representation from only one low-resolution sample.
\end{itemize}

\begin{itemize}
  \item The smallest size of the input low-resolution image (indicated in column 7).
  Some databases are naturally captured at a certain distance, as it can be observed in Figure~\ref{fig:dbs}.
  For example:
  CASIA Iris Mobile v2.0 (with images having an average iris diameter of 132 pixels), Q-FIRE (110 pixels),
  or the MBGC Portal database (90 pixels).
  %
  %
  To achieve a smaller image size, a number of studies perform sub-sampling of high-resolution images (indicated in column 8).
  Simulated downsampling is a common practice in 
  the super-resolution literature \cite{[Dong16]}, \cite{[Wang14]},
  due to the lack of databases with very low-resolution images.
  For example, in the work \cite{[Nguyen11a]}, the authors down-sampled images from the MBGC Iris video database to an average iris diameter of 50 pixels. Images of the same database were reduced to an average diameter of 20 pixels in the work \cite{[Jillela11]}.
    The CASIA Interval database has been also used for the same purpose in a number of studies \cite{[Shin09],[Alonso15b],[Alonso16b],[Ribeiro17]}. The average iris diameter of sub-sampled images in these studies ranged from 11 to 53 pixels.
    Finally, in our studies with the VSSIRIS database \cite{[Alonso17a],[Ribeiro18]}, we down-sampled iris images to an average diameter of 13 pixels.
    %
  %
\end{itemize}

\begin{itemize}
  \item The features used to carry out recognition experiments (indicated in column 9). Most studies only employ the popular Iris Code representation \cite{[Daugman04]}. Very few works compare two or more feature extraction methods. Among those, the present paper
  stands out as the only one employing six different comparators.
\end{itemize}

\begin{figure}[t]
     \centering
     \includegraphics[width=.48\textwidth]{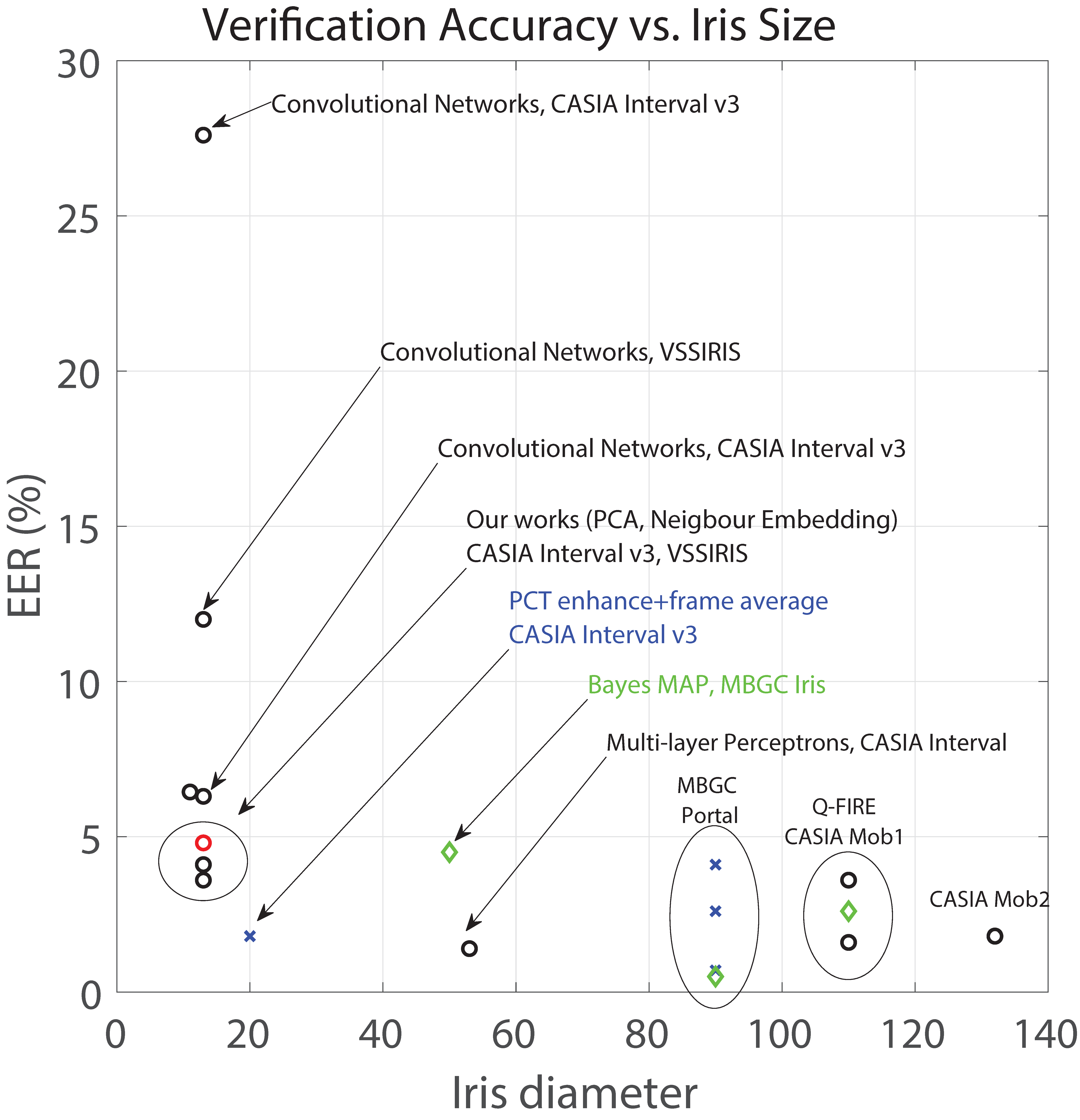}
     \caption{Verification accuracy reported by existing iris super-resolution studies vs. employed iris size (in pixels). The algorithm and/or database used is also mentioned.
     }
     \label{fig:LRsize_vs_EER}
\end{figure}

\begin{itemize}
  \item The biometric authentication results reported (indicated in columns 10-11) when images of the smallest size are used for recognition purposes. The present paper is the only one, together with \cite{[Huang03]}, which reports identification experiments.
  We further analyse these authentication results by plotting in Figure~\ref{fig:LRsize_vs_EER} the verification accuracy given in Table~\ref{tabla:SOA} against the iris size.
  Although the results are not directly comparable due to different databases and feature extraction methods being used, there is an inverse proportion between the EER and the diameter of the iris.
  Among the methods making use of very small iris images, our works with PCA
  \cite{[Alonso15b],[Alonso16b],[Alonso17a]}
  are among the most competitive.
  Recent studies adapting deep-learning frameworks \cite{[Ribeiro17],[Ribeiro18]} still report an accuracy significantly worse in some cases.
  There is one reconstruction-based method using PCT enhancement \cite{[Jillela11]} which also stands out for its excellent performance with an iris diameter of only 20 pixels.
  It is also worth noting that other works employing images with a higher diameter do not provide a significantly better performance, see for example the work \cite{[Shin09]}, based on Multi-layer Perceptrons, or the work \cite{[Nguyen11a]}, based on Bayes MAP probability estimation.
  The same appreciation can be done with the works employing the MBGC Portal, Q-FIRE or CASIA Mobile databases. Although these employ images with a iris diameter (in the range of 90-130 pixels), their performance is not much better in some cases.  
  These results suggest that there is still room for improving the performance of super-resolution methods in iris biometrics.

\end{itemize}


In the remaining of this section, we provide a brief description of the works summarized in Table~\ref{tabla:SOA}, categorized by the domain of operation (pixel or feature domain), and by the upsampling approach (reconstruction- or learning-based).

\subsection{Reconstruction-based methods in the pixel domain}

Reconstruction-based methods for iris images started in 2006 with the work of
Barnard \emph{et al}. \cite{[Barnard06]},
where they employed a multi-lens imaging hardware system to capture multiple iris images.
They carried out the reconstruction
by modelling the least square inverse problem associated with Equation~\ref{eq:acquisition-full}.
For this purpose, the blurring kernel, $B$, the warp function, $W$,
and the downsampling function, $D$, were estimated (the noise term was omitted).
To minimize the reconstruction error, they used the conjugate gradient method (CGLS).
In their experiments, they employed up to 15 low-resolution images as input.
They measured the quality of the reconstruction by computing the Hamming Distance between
the Iris Code of a reference high-resolution image, and the corresponding reconstructed image.
Experiments showed a reduction in the Hamming Distance in comparison with the distance to a low-resolution image.

Later in 2007, Fahmy \cite{[Fahmy07]} proposed an algorithm where high-resolution iris images
were estimated using an auto-regressive model to fuse a sequence of low-resolution iris images.
He first applied a cross-correlation model to register iris images from consecutive frames
of videos captured 3 feet away of the subject.
Then, an iris image which is 4-times higher in resolution was constructed from every 9
low resolution images.
This process was iterated until an image which is 16-times higher was obtained.
Two drawbacks of this method are that registration was done using the whole eye image,
and that they employed only focused images. These assumptions can be problematic
in unconstrained conditions, where off-angle or out-of-focus images may be present.
%

Video data from the Multiple Biometric Grand Challenge database \cite{[Phillips09]}
has been used in several studies \cite{[Hollingsworth09a],[Nguyen10a],[Nguyen10],[Nguyen11]},
where several polar images were aligned and combined pixel-wise
to obtain a reconstructed image according to Equation~\ref{eq:reconstruction_fuseim}.
%
%
In the work \cite{[Hollingsworth09a]}, Hollingsworth \emph{et al}. created a single average polar image
from 10 multiple frames of a frontal iris video.
Data employed was from the MBGC Iris NIR database, 
%
having an average iris diameter of 220 pixels.
The ten best focused images from each video were selected
by employing the filter kernel proposed in \cite{[Kang07]}.
They tested both the mean and median fusion,
concluding that the mean is better, since it employs statistics from all available pixels.
The median, on the other hand, only employs statistics of one or two pixels.
They achieved an EER of 0.7\% by fusion of 10 frames, compared to an EER of 1.56\% obtained by employing only one-gallery and one-probe frame.
In the work \cite{[Nguyen10a]}, Nguyen \emph{et al}. employed the robust mean,
which consist of fusing intensity values of an individual pixel over multiple frames
by taking the mean of values within two standard deviation from the mean.
Data employed was from the MBGC Portal NIR database, 
%
with an average iris diameter of 90 pixels.
Authentication experiments were done by comparing super-resolved images to
high resolution still iris images captured with a close-up NIR camera,
which are also provided with the MBGC Portal dataset (Figure~\ref{fig:dbs}).
The authors argued that in unconstrained environments like the MBGC Portal database,
extreme pixel values can appear in different locations in different frames
due to reflections, shadows, eyelids, or eyelashes.
For this reason, the proposed approach is more robust against unexpected extreme pixel values.
The obtained EER is of 4.1\%, contrasted to 4.6\% when the approach
of \cite{[Hollingsworth09a]} was applied.

Considering that frames in an iris video sequence can have different quality when
captured in adverse conditions, Nguyen \emph{et al}. \cite{[Nguyen10],[Nguyen11]}
employed quality measures to compute the weights of Equation~\ref{eq:reconstruction_fuseim}.
%
In the paper \cite{[Nguyen10]}, they employed the focus level of each frame,
which was measured by evaluating the high frequency total energy of the image.
A high-resolution image was then estimated using the focus-score weighted average
of the available frames.
Heavily de-focused frames were discarded from further processing,
while the others were fused to super-resolve the iris image.
With the proposed approach, they achieved an EER of 2.1\% using the MBGC Portal database.
In the paper \cite{[Nguyen11]}, the authors combined four quality factors
(focus, off-angle, motion blur, and illumination variation) into a unified
quality score for each iris frame. For this purpose, they employed the
Depmster-Shafer theory proposed in \cite{[Kalka10]}.
Another novelty of this work was that instead of using the conventional weighted average of Equation~\ref{eq:reconstruction_fuseim},
frames were fused by using an exponential weighted average.
Experiments were also carried out to determine the optimum number of frames for fusion.
The authors concluded that 5 is the optimal number of frames to fuse, although they acknowledged that
a different dataset may lead to a different number if the acquisition conditions are different.
The reported EER in this case is of 0.7\%.
When the number of frames increases
beyond 5, they observed that the poor quality of the additional frames counteracted
the introduction of extra information.
Following a similar vein, Othman \emph{et al}. \cite{[Othman13],[Othman15]} expanded this idea
by computing the quality of local image patches.
For this purpose, they estimated a Gaussian Mixture Model (GMM) of clean iris texture distribution.
Then, during the fusion, each pixel was weighted individually with the quality value
of the associated local patch, instead of employing a single quality score for the whole image.
This way, regions with better quality contribute more in the reconstruction of the fused image
than regions with poorer quality.
Using the MBGC Portal database, they reported an EER of 2.58\%,
compared to an EER of 4.9\% obtained by simple pixel intensity average.
They also employed the Q-FIRE database, which contains videos
captured at 5, 7, and 11 feet away with a telephoto lens.
%
%
The number of input images employed for reconstruction varied from 2 to 10,
with the frames ranked according to their quality. 
In the experiments, the authors observed that performance with Q-FIRE improved until the
best 6 frames were chosen, then the error increased when extra frames were added.
The reported FRR @ FAR=0.1\% with this database is of 2.54\% (images captured 5 feet away),
4.37\% (7 feet), and 2.04\% (11 feet).

%
Some works have included specific preprocessing for image enhancement purposes.
For example, Jillela \emph{et al}. \cite{[Jillela11]} applied
Principal Component Transform (PCT, variation of PCA) to polar iris images
of the MBGC Iris database
in order
to highlight the variance information among the pixel intensity.
Then, the enhanced images were fused by image averaging.
The optimum number of frames per video was empirically chosen as 6,
with the best quality frames selected manually.
Low-resolution data was generated by sub-sampling the original images
by a factor of 1/2, 1/4, and 1/8.
This resulted in low-resolution image sets 
with an average iris diameter of 
110, 50, and 20 pixels, respectively.
Authentication experiments were done by comparing super-resolved images to
a separate set of original high-resolution images 
set aside as gallery set.
The reported EER using images with the lowest resolution is of 1.76\%.
When no reconstruction is carried out (i.e. each low-resolution frame is
compared separately against gallery frames), the reported EER is of 6.09\%,
highlighting the benefit of the proposed approach.
In the work by Hsieh \emph{et al}. \cite{[Hsieh16]} the authors incorporated
optical wavefront coding techniques \cite{[Dowski95]} to increase the depth of
field (DoF) in long-range iris portal acquisition.
This was achieved by optimizing the optical architecture of the acquisition hardware.
An extended depth of field (EDoF) allows a higher capture volume as the subjects walks
to the camera.
They also exploited image quality measures to weight the contribution of low-resolution
images by exponential weighted average.
They employed their own video data from 16 subjects captured with a telephoto lens.
Enrolment images were captured with subjects standing at 3m, and test images were
captured 
at 11 defocus positions
(from -15 to +15 cm in 3 cm intervals).
For each defocus position, two images were captured, one without the EDoF hardware,
and one with the EDoF hardware.
Quality of each frame was assessed by calculating the Hamming distance between all
possible pairs of the 11 test images, and then computing the average distance
of all pair-wise comparisons associated to each test image. A high quality image
is expected to have a lower average value, and vice versa.
A novelty of this approach
with respect to the already mentioned approaches of this section
is that reconstruction is made in local patches.
When the 11 test images captured with the EDoF hardware are fused following the proposed method,
the reported recognition results are EER=0\%, and FFR=0\% at a FAR of 0.1\%.
On the contrary,
when the images are captured without the EDoF hardware, the paper reports
an EER of 11.5\%, and a FRR of 37\% at a FAR of 0.1\%.

The CASIA Long Range database has also been used in a number of studies \cite{[Deshpande17a],[Deshpande17]}.
%
Deshpande and Patavardhan \cite{[Deshpande17a]} employed
Gaussian Process Regression (GPR) and Enhanced Iterated Back Projection (EIBP)
to super-resolve iris images.
The best frame was selected as reference for alignment purposes by using the Discrete Cosine Transform.
To account for local image deformations, they carried out reconstruction in local patches of the polar image.
A threshold is applied to the intensity variance of each patch.
If the variance is higher than the threshold, the patch is reconstructed with GPR,
otherwise it is reconstructed with EIBP.
The GPR is a time consuming process, therefore patches with less amount of information
(measured by their variance of intensity) are processed with the faster EIBP algorithm.
Performance was evaluated in \cite{[Deshpande17a]} by downsampling iris images, and then super-resolving them.
The authors reported several image fidelity measures in the pixel domain between
original high-quality images and reconstructed images.
Recognition results were also reported, with a FAR of 3.86\% and a FRR of 4.21\%.
although no information is given regarding the size of the low-resolution images
involved in the authentication experiments.
The same authors \cite{[Deshpande17]} employed an enhanced
Total Variation regularization algorithm \cite{[Ng07]} to super-resolve iris images.
Low resolution input images were first de-blurred in order to remove
motion blur and then, motion estimation between consecutive image frames was computed.
In the regularization process, the estimated blur kernel and motion vectors were taken
into account to iteratively generate a high resolution reconstructed image.
The authors employed six polar images of 300$\times$40 pixels to estimate one super-resolved image of twice the input resolution (600$\times$80 pixels).
The authors evaluated the algorithms by reporting several image fidelity and textural measures between
original high-quality images and reconstructed images.
However, no person authentication experiments were reported.
Iterative back projection was also employed by Ren \emph{et al}. \cite{[Ren12]} in a previous work.
A frame was randomly selected as reference for alignment, which was done in the Fourier domain.
The output high-resolution image was initialized by upsampling the first low-resolution frame
using nearest neighbour interpolation.
Then, the estimated high-resolution image was iteratively updated with the gradient of the total square error
in the pixel domain between the reference image and the low-resolution frames.
When the total square error achieved a threshold value, the iterative process was finished.
The authors employed iris images from the CASIA 3.0 database in their experiments.
Low-resolution data was generated by sub-sampling polar images of 512$\times$64 pixels
to reduce their size by a factor of 1/2, and 1/4.
The resulting polar images were of size 256$\times$32, and 128$\times$16 pixels, respectively.
The EER reported by the authors without reconstruction is of 12.75\% (with polar images of 256$\times$32  pixels)
and 13.7\% (polar images of 128$\times$16).
Sub-sampled polar images were then reconstructed to their original size of 512$\times$64.
The number of input images evaluated for reconstruction was 2, 4 and 6,
concluding that 4 images is a good compromise between performance and processing time.
With the proposed algorithm, the reported EER is of
6.87\% and 8.69\% (using polar images of 256$\times$32 and 128$\times$16 pixels, respectively).

\subsection{Learning-based methods in the pixel domain}
\label{subsec:learning-based-pixel}

Learning-based iris reconstruction was first proposed in 2003 by Huang \emph{et al}. \cite{[Huang03]}.
In this method, the probabilistic relation between different frequency bands is learned, in order to predict the missing high-frequency information of low-resolution images.
It is based on the general purpose method by Freeman \emph{et al}. \cite{[Freeman02]}.
The training set of high-resolution polar images is first pre-processed as follows.
Each high-resolution image is separated in three bands:
low-frequency, by downsampling and upsampling the high-resolution image;
medium-frequency, by applying a Circular Symmetric Filter (CSF);
and high-frequency, by subtracting the high-resolution and the low-frequency images.
Images are then divided into patches,
which each position having associated a set of low-, medium-, and high-frequency patches.
Given an input low-resolution image to be reconstructed,
it is first up-sampled by cubic interpolation.
Then, a feature image is constructed by applying a Circular Symmetric Filter (CSF)
to extract the medium-frequency information.
The image is then divided into patches. For each patch, the 200 patch sets
of the training set whose feature vectors are closest to the input patch are selected
using the L1 distance.
The best matching set from this sub-set is then computed
based on spatial constraints at adjacent patch borders,
and the corresponding high-frequency patch is selected.
A high-frequency image is then obtained by stitching together
the resulting high-frequency patches.
The output reconstructed image is finally obtained by adding the high-frequency
image to the test input image.
In the experiments reported in this paper,
low resolution data was generated by sub-sampling images
from the CASIA dataset to three different low-resolutions
(not specified).
The experiments report a significant improvement
in rank-1 and EER metrics in comparison with cubic interpolation,
and with the original method described in \cite{[Freeman02]}.

In the work \cite{[Shin09]}, Shin \emph{et al}. employed
multiple Multi-layer Perceptrons (MLP) to restore local iris patches.
Each block of the input image is classified into one of 3 types
(vertical, horizontal, and non-edge) based on difference of pixel intensities.
Three MLPs are trained, one per type of block, to estimate
selected pixel values of the high-resolution patch.
An advantage of this method is that it does not require
accurate image registration.
Reconstructed blocks are then assembled together, and
missing pixels are filled by bilinear interpolation.
This is because the MLPs are not trained to predict all
pixels of the high-resolution patch, but only a part of them.
Low-resolution data was generated in \cite{[Shin09]} by sub-sampling images
from the CASIA Interval v3 database to 1/3 and 1/4 of the original image size.
This resulted in image sets 
with an average iris diameter of 70 and 53 pixels.
The MLPs were trained using 12 randomly selected images from different eyes.
The reported EER using the smallest images is of 1.39\%,
compared with 1.49\% by bilinear interpolation,
or 0.89\% with original high-resolution images.

Sparse representation in over-complete dictionaries was used in the work of Aljadaany \emph{et al}. \cite{[Aljadaany15]}.
Traditional approaches in this regard, such as K-Singular Value Decomposition (K-SVD),
have the limitation that the number of dictionary items
and the number of sparse coefficients has to be predefined.
To overcome this limitation, the authors used a non-parametric Bayesian approach, named Beta Process (BP),
to build the discriminative over-complete dictionary and discover the necessary parameters automatically.
During the training phase, high-resolution polar images are down-sampled to create their
low-resolution counterparts, which are then used to learn the relationship
between high- and low-resolution patches.
The dictionaries in both manifolds are assumed to have the same sparse weights,
which simplifies the reconstruction process.
During the testing phase, sparse weighs of the low-resolution image are first computed.
Then, the weights are transferred to the high-resolution manifold,
which are then used to calculate the conditional expectation of a high resolution iris image.
This approach was evaluated in a subset of the CASIA database.
Low-resolution images were generated by sub-sampling images
to 25\% of their original size and adding Gaussian noise.
The authors reported and improved recognition performance with the proposed method (EER $\sim$2\%),
in comparison with linear interpolation (EER $\sim$3\%).

Instead of synthesizing high-resolution iris images, Liu \emph{et al}. \cite{[Liu14]}
learned a non-linear mapping function in which
homogeneous (high resolution-high resolution)
and heterogeneous (high resolution-low resolution) comparison scores
are non-linearly mapped into a common high dimensional space.
In this space,
separation between inter- and intra-class distributions are maximized
regardless whether they originate from homogeneous or heterogeneous samples.
This converts the multi-class problem (discriminating between different identities)
into a two-class problem (discriminating between inter- and intra-class comparisons).
Each class is composed of two sets: homogeneous comparisons (high-to-high resolution)
and heterogeneous comparisons (high-to-low resolution).
During the training phase,
all possible inter/intra-class comparisons of training data are carried out.
Polar images are divided into patches,
and comparisons are done for each patch separately,
so given two images, their comparison results in a set of patch-scores.
During the testing phase, a low-resolution test sample is compared
to the gallery of high-resolution templates, and the resulting
scores are mapped to the learned common space,
where a rejection/acceptance decision can be made.
The authors employed the Q-FIRE database.
Data captured at 5 and 11 feet away were selected respectively as high-
and low-resolution sets.
Eye detection and iris segmentation were carried out in the input iris
videos, resulting in multiple images available for each subject.
Heavily de-focused or occluded images were discarded for further use.
A total of 1400 images (700 low- and 700 high-resolution images)
from 100 different classes were used as training data.
The reported EER with the proposed method is of 1.6$\pm$0.92\% (10-fold validation),
compared with $\sim$4.6\% without score mapping.

In the work \cite{[Alonso15b]},
we presented an iris reconstruction method based on Principal Component Analysis (PCA)
that is the basis of the present paper.
The technique is inspired by the system of \cite{[Chen14]} for face images.
Given a test low-resolution image,
a PCA Eigen-transformation is conducted for each patch using
a set of low-resolution training images.
After the reconstruction weights are computed in the low-resolution manifold,
they are transferred to the high-resolution manifold,
where high-resolution patches are then reconstructed
using a linear combination of collocated high-resolution patches
from training images.
The method was evaluated using the CASIA Interval v3 database, and
Log-Gabor wavelets \cite{[MasekThesis03]} as feature extraction method.
Low-resolution data was generated by sub-sampling high-resolution images
to reduce their size by a factor from 1/2 to 1/18
(the latter corresponding to 
an average iris diameter of 11 pixels).
%
The paper reported an EER of 6.44\% with images of the lowest resolution,
compared with 12.23\% when bi-cubic reconstruction is used.

The current paper extends our previous studies with the PCA method,
including additional comparators and experiments.
A limitation of this method is that it assumes that low- and
high-resolution manifolds have similar local geometrical structure.
Reconstruction weights are estimated on the low-resolution manifold,
and they are simply transferred to the high-resolution manifold.
However, the geometrical structure
of the low-resolution manifold is distorted by the one-to-many relationship between low- and
high-resolution patches \cite{[Li09]}.
Therefore, the reconstruction weights estimated
on the low-resolution manifold do not necessarily correlate with the actual weights needed to
reconstruct the unknown high-resolution patch.
To cope with this limitation, we later considered to use iterative neighbour embedding of
local patches (LINE) \cite{[Alonso17]}, where the geometry of the low- and high-reconstruction
manifolds are jointly taken into account during the reconstruction.
During the testing phase, the reconstruction weights are computed by minimizing
a regularization function that considers the distance of the input patch
to the training dictionary both in the low- and in the high-resolution manifolds.
The lowest resolution evaluated
consisted of images with
an average iris diameter of 13 pixels.
The LINE method compared well with the PCA method,
showing additional performance improvements at very low resolutions.
The PCA and LINE methods were also evaluated in \cite{[Alonso17a]} using
smart-phone images from the VSSIRIS database \cite{[Raja14b]}.
In this work,
high-resolution images
were down-scaled by a factor of 1/22
(corresponding to 
an average iris diameter of $\sim$13 pixels).
The experiments showed a superior performance of the trained
reconstruction approaches in comparison to bilinear
or bicubic methods,
with the LINE approach showing better performance than PCA.
The best recognition rates reported were 4.64\% (PCA) and 4.1\% (LINE) with the fusion of
log-Gabor wavelets and the SIFT operator.

Recent studies have also adapted deep-learning frameworks
to the task of iris super-resolution \cite{[Zhang16a],[Ribeiro17],[Ribeiro18]}.
Zhang \emph{et al}. \cite{[Zhang16a]}
adapted
Super-Resolution Convolutional Neural Networks (SRCNN) \cite{[Dong14]}
and Super-Resolution Forests (SRF) \cite{[Schulter15]}
to reconstruct iris images in the polar domain.
The SRCNN employed learns the non-linear mapping
function between low- and high-resolution images with 3 layers:
the first one extract feature maps of low-resolution patches,
the second one maps these feature vectors into feature maps of corresponding high-resolution patches,
and the last one aggregates high-resolution patches to generate the output  image.
The loss function employed is the mean squared error between the reconstructed
images and the corresponding ground-truth high-resolution image, which in turn
corresponds to a fidelity measure between images, and not to a performance metric.
This is common to all iris super-resolution studies that employ deep-learning frameworks, which may explain that their performance is still behind methods not based on deep-learning.
In the SRF method, Random Forest are used to directly map
low-resolution patches to high-resolution patches.
During tree growing, a regularized objective function
that operates on both output and input domains is used,
so higher quality results can be achieved.
Training of SRCNN and SRF was done with 91 non-iris images,
of use in other super-resolution studies.
The two methods were tested using the CASIA Mobile v1.0 and v2.0 databases, which
contain near-infrared (NIR) images captured with smart-phones
by using a NIR imaging module. 
%
The EER achieved with CASIA Mobile v1.0 is of 3.65\% (SRCNN) and 3.61\% (SRF), compared
with 3.89\% obtained using iris images without enhancement. These
results corresponds to the score fusion of left and right iris.
%
With CASIA Mobile v2.0, the authors reported single-eye experiments comparing images captured at various distances, namely 20-20 cm, 20-25 cm, and 20-30 cm.
Reported experiments show that applying the SRCNN and SRF methods results
in better performance in comparison with employing images without enhancement, specially at
low FAR.

In the works by Ribeiro \emph{et al}. \cite{[Ribeiro17],[Ribeiro18]},
the authors employed several deep learning methods
to reconstruct iris images. 
In these works,
images were divided into patches,
which were then reconstructed separately using each respective network
to obtain high-resolution patches.
The work \cite{[Ribeiro17]} employed
Super-Resolution Convolutional Neural Networks (SRCNN) \cite{[Dong14]}
and Stacked Auto-Encoders (SAE) \cite{[Bengio13]}.
The experimental framework used the CASIA Interval v3 database
as in the work \cite{[Alonso15b]}.
High-resolution images
were down-scaled by a factor from 1/2 to 1/16
(the latter corresponding to 
an average iris diameter of 13 pixels).
Using Log-Gabor wavelets, the PCA method of \cite{[Alonso15b]} still shows better
performance at the lowest resolution (EER of 4.79\% vs. 6.26\%).
However, the paper evaluated a second comparator based on local SIFT key-points \cite{[Lowe04]},
with which the deep-learning methods showed better recognition performance
(EER of 19.5\% with PCA vs. 17.26\% with SRCNN).
%
In the work \cite{[Ribeiro18]}, the authors evaluated
Super-Resolution Convolutional Neural Networks (SRCNN) \cite{[Dong14]},
Very Deep Convolutional Neural Networks (VDCNN) inspired by VGG-net \cite{[Kim15Corr]},
and Super-Resolution Generative Adversarial Networks (SRGAN) \cite{[Ledig16Corr]}.
Besides the CASIA Interval v3 database,
the authors also employed smart-phone images from the VSSIRIS database
in their experiments.
As in \cite{[Ribeiro17]},
images of both databases were down-sampled up to an average iris diameter
of 13 pixels.
They also explored the use of different databases to
train the networks for the iris reconstruction task.
The databases employed include texture databases, natural image databases,
and iris databases. The iris databases contain both
near-infrared and visible wavelength images,
while the other two categories only include visible data.
An interesting finding of this paper is that using texture databases
or iris images from other databases
for training provides good reconstruction results,
even if captured with different lightning.
%
At the lowest resolution, 
the best recognition performance with CASIA images was given by SRCNN,
with a reported EER of 27.6\%. 
%
With the VSSIRIS database,
the best performance at the lowest resolution 
was given by SRGAN, 
with an EER of 12\%. 
It should be mentioned nevertheless that the recognition features
used in \cite{[Ribeiro18]}
are different than those in previous studies
with the same databases
\cite{[Ribeiro17],[Alonso17a]},
which might explain the differences in performance.

\subsection{Learning-based methods in the feature domain}
\label{subsec:learning-based-feature}

Instead of super-resolving pixel intensity values,
the methods in this section super-resolve images in the feature space.
A commonality found in most of them
is that they employ several input images, so they
could be considered to be reconstruction-based.
However, we categorize them as learning-based since the mapping
relation between low- and high-resolution images
is learned using training dictionaries.
%
%
%

In the work published in 2011, Nguyen \emph{et al}. \cite{[Nguyen11a]}
carried out the reconstruction by modelling the inverse problem
associated with Equation~\ref{eq:acquisition-full}.
However,
they replaced the low- and high-resolution images $\overline X$ and $\overline Y$
with their PCA feature representations \cite{[Turk91]}
estimated from gallery images in the polar domain.
Given a low-resolution test image, it is first projected onto the PCA space.
Then, a high-resolution iris image is estimated by Bayes MAP probability estimation,
which is computed using iterative steepest descent.
The method was tested with data from the MBGC Iris NIR database.
Two high quality iris images were selected per subject, one used as gallery,
while the other was degraded by Gaussian blurring,
random warping and downsampling by
a factor of 1/4 to create a series of 16 low resolution images.
The average iris diameter of low-resolution images in this case
was of 50 pixels.
In addition, the method was compared with the quality-weighted
pixel average technique of the same authors,
published in \cite{[Nguyen10]}.
The effectiveness of the reconstruction was measured
by computing the distance of the reconstructed feature vector
to the true feature vector of the original high-resolution image.
Reported experiments showed that the proposed feature-domain
approach produces closer features than the method in \cite{[Nguyen10]},
as well as better recognition performance.
The proposed feature-domain approach achieved an EER of 4.5\%,
compared to an EER of 10\% obtained when using the pixel average method.

Nguyen \emph{et al}. identified as a drawback of the previous approach
that linear features such as PCA are not optimal for recognition,
when compared to nonlinear ones
such as iris codes extracted from Gabor phase-quadrant encoding \cite{[Daugman04]}.
Since Gabor-based features
are shown to be one of the most discriminant features
for face \cite{[Serrano10]} and iris \cite{[Daugman07]},
they proposed to super-resolve iris images in the Gabor domain \cite{[Nguyen12],[Nguyen13]}.
A challenge of this approach comes from the difficulty
of modelling the relationship
between low- and high-resolution images in
non-linear feature domains.
However, they observed that the response to a Gabor wavelet is linear,
whilst the nonlinearity of the process comes
from the phase-quadrant encoding applied to compute the iris code.
Thus, following the same strategy of \cite{[Nguyen11a]},
they replaced
the low- and high-resolution images $\overline X$ and $\overline Y$
of Equation~\ref{eq:acquisition-full}
with
the complex-valued responses of polar images to Gabor wavelets.
%
In this occasion, the method was tested with data
from the MBGC Portal NIR database.
Four video sequences of each identity
were matched against still high-resolution images.
For each video, the frames with the best quality were selected
according to the Depmster-Shafer quality assessment of \cite{[Kalka10]}.
A threshold to remove poor quality frames was selected
through experimentation.
With the proposed approach, they achieved an EER of 0.5\%.
The method outperformed other approaches evaluated in the paper,
including
bicubic (EER=1.8\%),
pixel average \cite{[Hollingsworth09a]} (EER=1.4\%),
quality-weighted pixel average \cite{[Nguyen10]} (EER=0.9\%),
PCA feature representation \cite{[Nguyen11a]} (EER=4.8\%),
or
LDA feature representation (EER=2.1\%).

Lastly, in the work \cite{[Liu13]}, Liu \emph{et al}. learnt the statistical
relationship between a set of binary codes from low-resolution images,
and the binary code of the corresponding high-resolution image.
For this purpose, they employed Markov networks.
The co-occurrence of neighbouring bits in the high-resolution iris code
was also modelled with the network.
Besides the non-linear relationship between feature codes
of low- and high-resolution iris images,
the Markov model is also able to produce a
weight mask which measures the reliability of
each bit in the enhanced iris code.
This weight mask can be used in the computation
of the Hamming Distance between two iris codes
to further
enhance recognition accuracy.
The authors employed the Q-FIRE database.
Images captured at 5 and 11 feet away were selected respectively as high-
and low-resolution sets.
Eye detection and iris segmentation was carried out on the input iris
videos, resulting in multiple iris for each subject.
Heavily de-focused or occluded images were discarded for further use.
A total of 1000 images (500 low- and 500 high-resolution images)
were used as training data.
The reported EER with the proposed method is of 2.6\%.
Reported experiments also show that it outperforms
other existing algorithms, including \cite{[Hollingsworth09a],[Nguyen11],[Nguyen12]}, specially at low FAR.
In a later work with the same database \cite{[Liu14]} (presented in Section~\ref{subsec:learning-based-pixel}),
the same authors improved the EER further to 1.6$\pm$0.92\%.

\begin{figure*}[htb]
     \centering
     \includegraphics[width=.95\textwidth]{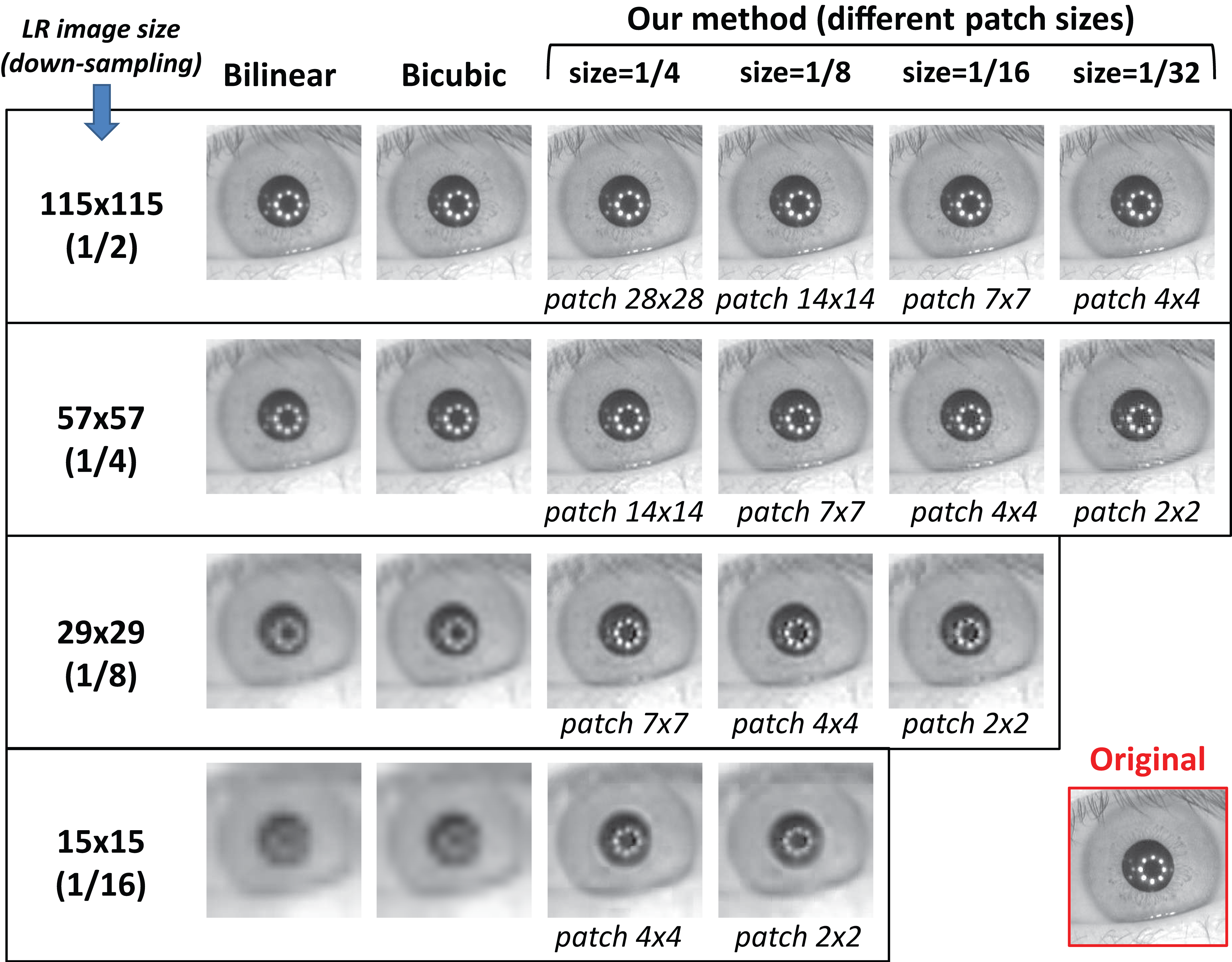}
     \caption{Resulting high-resolution hallucinated images for different downsampling factors
and patch sizes. 
}
     \label{fig:images-example}
\end{figure*}

\section{Eigen-Patch Iris Super-Resolution}
\label{sect:method}




The block diagram of the proposed iris super-resolution method is shown in
Figure~\ref{fig:system}, which is based on the eigen-patch
hallucination method for face images proposed by Chen and Chien~\cite{[Chen14]}.
Each input image $\overline X$ is first separated into overlapping patches.
An eigen-transformation is then conducted on each patch using collocated
patches of low-resolution iris images from a training set $\{\bar L\}$, in order
to obtain the optimal reconstruction weights of each patch.
The reconstruction weights are then transferred to the high-resolution manifold,
where the high-resolution patch is rendered using collocated patches of the
high-resolution images in the training set $\{\bar H\}$.
A preliminary high-resolution image $\tilde Y '$ is then formed
by stitching the reconstructed high-resolution patches.
Finally, a reprojection operation is applied to further reduce artefacts
and make the output high-resolution image $\tilde Y$ more similar to the input low-resolution image.
The methods is described in more detail in the following sub-sections.

\subsection{Eigen-Patches}
\label{subsec:eigenpatches}
Without loss of generality, suppose that our image recognition problem is iris recognition.
Given an input low-resolution iris image $\overline X$, it is
first separated into $N = N_v \times N_h$ overlapping patches $\left\{
{\overline x } \right\} = \left\{ {\overline {x_1 } ,\overline {x_2
} , \cdots ,\overline {x_N } } \right\} $, where $N_v$ and $N_h$ are
the vertical and horizontal number of patches, respectively. Since
we will consider square images in our experiments, we can assume
from the remainder of this paper that $N_v = N_h$.

Two super sets of basis patches are computed for each patch position $i$ from
collocated patches of a training database of high resolution images $\{\bar H\}$.
One of the super sets is obtained from collocated high resolution patches as follows.
For each patch position $i$, we stack patches at the same position from the set of $M$ high-resolution training iris images as shown in Figure~\ref{fig:dictionary}, which we denote as $\bar {H}_i = \left\{ {\overline {h_i^1 } ,\overline {h_i^2 }
, \cdots ,\overline {h_i^M } } \right\} $.
By degradation (low-pass filtering and downsampling) using the acquisition model defined in Equation~\ref{eq:acquisition}, a low-resolution database $\{\bar L\}$ is generated from $\{\bar H\}$, and
%
the corresponding low-resolution super set
$\bar {L}_i = \left\{ {\overline
{l_i^1 } ,\overline {l_i^2 } , \cdots ,\overline {l_i^M } } \right \} $ is obtained for each patch position $i$.
This way, the structure of the iris image is exploited by the construction of position-dependent dictionaries.
In the example in Figure~\ref{fig:dictionary}, the high-resolution
dictionary $\bar {H}_1$ (marked in red) is composed using the top-left position patches
of all M high-resolution images, while the corresponding low-resolution dictionary
$\bar {L}_1$ is composed using the corresponding top-left patches of the M low-resolution images.

During testing, a PCA Eigen-transformation
is conducted for each low-resolution input patch $\overline x _i$ using the collocated patches of the low-resolution dictionary $\bar {L}_i $
to compute the optimal linear reconstruction weights $\overline {c_i
}  = \left\{ {c_i^1 ,c_i^2 , \cdots ,c_i^M } \right\}$. More specifically, we first compute the mean-patch of the $i$-th low-resolution dictionary using

\begin{equation}
\overline{m}_i^L = \frac{1}{M} \sum_{j=1}^{M}{\overline{l}_i^j}
\label{eq:mil}
\end{equation}

\noindent so that the low-resolution dictionary $\bar {L}_i$ for the $i$-th patch is centred by removing the mean-patch $\overline{m}_i^L$:

\begin{equation}
\bar L _i^{'} = \bar {L}_i - \overline{m}_i^L
\end{equation}

For an input patch $\overline{x}_i$, the weight vector can be computed by projecting it onto the eigen-space using

\begin{equation}
\overline{w}_i = E_i^T \left ( \overline{x}_i - \overline{m}_i^L\right)
\end{equation}

\noindent where the eigen-patches $E_i$ are derived by applying PCA to the covariance matrix $\bar {C}_i$ of the centred dictionary $\bar L _i^{'}$:

\begin{equation}
\bar{C_i} = \bar L _i^{'T} \bar L {_i^{'}}
\end{equation}

The matrix $E_i$ can be decomposed using

\begin{equation}
E_i = \bar{L}_i^{'} V_i {\Lambda_i}^{-\frac{1}{2}}
\end{equation}

\noindent where $V_i$ and $\Lambda_i$ are the eigenvectors and eigenvalues provided using PCA.
In this work, we retain the eigenvectors of $\bar {C}_i$ which contains at least 99\% of the variance.
The reconstruction weights are then derived using

\begin{equation}
\overline{c}_i = V_i {\Lambda_i}^{-\frac{1}{2}}\overline{w}_i
\end{equation}

The $i$-th high-resolution patch $\tilde{y}_i$ is then reconstructed from the collocated patches of the high-resolution dictionary $\bar {H}_i $ using

\begin{equation}
\tilde{y}_i = \sum_{j=1}^{M}{c_i^j \overline{h}_i^j} + \overline{m}_i^H
\end{equation}

\noindent where $\overline{m}_i^H$ is the high-resolution counterpart of Equation~\ref{eq:mil}:

\begin{equation}
\overline{m}_i^H = \frac{1}{M} \sum_{j=1}^{M}{\overline{h}_i^j}
\label{eq:mih}
\end{equation}

The recovered patches are then stitched together by averaging overlapping pixels to synthesize the preliminary high-resolution iris $\tilde Y '$.
It is important to mention here that every patch, which represents a particular spatial region within the iris, is optimized using dictionaries of irises at the same position.
Therefore, the different reconstruction weights are optimized for every region of the iris. This ensures that iris images of higher quality, which are locally optimized, are reconstructed.



\subsection{Image Reprojection}
A re-projection step is further applied to $\tilde Y '$ to reduce
artefacts and make the output image $\tilde Y$ more similar to
the input image $\overline X$. The image $\tilde Y '$ is
re-projected to $\overline X$ using the model of  Equation~\ref{eq:acquisition} via:

\begin{equation}
\tilde Y ^{t + 1}  = \tilde Y ^t  - \tau U\left( {B\left(
{DB\tilde Y ^t  - \overline X } \right)} \right)
\end{equation}

\noindent where $U$ is the upsampling matrix. The process stops when
$|\tilde Y ^{t + 1}-\tilde Y ^{t}|$ is smaller than a
threshold. For our experiments in iris biometrics,
we use $\tau$=0.02 and $10^{ - 5}$ as the difference
threshold.

\section{Evaluation Framework}
\label{sect:framework}

\subsection{Dataset}
\label{sect:db}

For our
experiments,
we used the CASIA Interval v3 iris database \cite{[CASIAdb]}.
It consists of 2,655 NIR images from 249 contributors,
captured in 2 sessions (the number of images per
contributor and per session is not constant). A close-up iris camera
was used to capture the images, with a resolution of 280$\times$320
pixels. Manual annotation of
the database is available \cite{[Alonso15],[Hofbauer14]}.
All images were resized via
bicubic interpolation to have the same sclera radius ($R$=105,
average sclera radius of the whole database according to the
ground-truth). Then, images were aligned by extracting a square region
of 231$\times$231 pixels around the pupil centre, which corresponds
to about
1.1$\times R$.
In case that such extraction was not possible (for
example if the eye is close to an image side), the image was
discarded. After this procedure, 1,872 images remained, which were
used for our experiments.
The dataset of aligned images was further divided into two sets: a
training set comprised of images from the first 116 contributors
($M$=925 images), used as dictionary images to train the eigen-patch hallucination method,
and a test set comprised of the remaining 133 contributors (947
images), which was used for validation.

The test and dictionary images were
%
down-sampled via bicubic interpolation using MATLAB's \texttt{imresize} function
by 
$1/n$, with $n \in \left\{ {2,4,6,8,10,12,14,16} \right\}$.
%
This resulted in down-sampled images of
115$\times$115, 57$\times$57, 39$\times$39, 29$\times$29, 23$\times$23,
19$\times$19, 17$\times$17, and 15$\times$15 pixels respectively.
The dictionaries for each patch and for each downsampling factor
were constructed using the position-patch method
described in Section~\ref{subsec:eigenpatches}.
Down-sampled test images were then
used as input low-resolution images, from which hallucinated
high-resolution images were computed using the proposed algorithm.
This simulated downsampling is the approach followed in most
previous studies \cite{[Wang14]}, due to the
lack of databases with low-resolution and corresponding
high-resolution reference images.
The proposed method was compared with bilinear and bicubic interpolation.
The method were implemented in MATLAB.
All simulations were run using a machine with Intel (R) Core (TM) i7-4600U CPU at
2.10GHz running Windows 64-bit Operating system.

\begin{figure}[t]
     \centering
     \includegraphics[width=.45\textwidth]{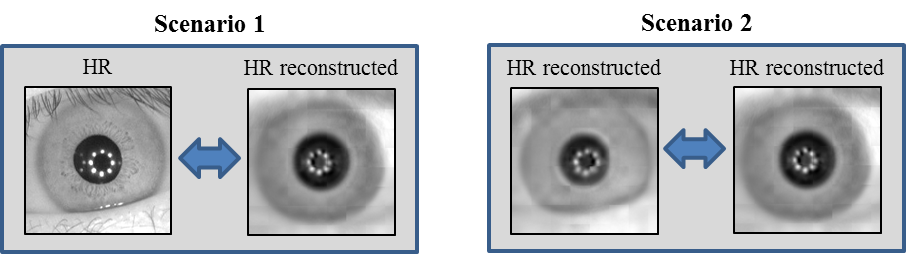}
     \caption{Scenarios considered in our identity recognition experiments.}
     \label{fig:scenarios}
\end{figure}

\subsection{Iris Authentication Experiments}
\label{sect:protocol}

We conducted verification and identification experiments in the test set with several
iris recognition algorithms (Section~\ref{sec:matchers}).
%
%
We considered two scenarios, shown in Figure~\ref{fig:scenarios}:
scenario 1), where enrolment samples were taken from original
high-resolution images of the test set, and query samples from hallucinated images;
and scenario 2), where both enrolment and query samples were hallucinated images.
The first case simulates a controlled enrolment scenario, where one of the samples
is of high-resolution. The second case, on the other hand,
simulates a totally uncontrolled scenario, where both samples are of low-resolution
(albeit for simplicity, both have similar resolution).

In our authentication experiments, each eye were considered
as a different user of the system.
\emph{Verification experiments} were done as follows.
Genuine trials were obtained by comparing each image of a user to the
remaining images of the same user, avoiding symmetric comparisons.
Impostor trials were obtained by comparing the $1^{st}$ image of a
user to the $2^{nd}$ image of the remaining users. With this
procedure, we obtained 2,607 genuine and 19,537 impostor scores
per comparator and per scenario.
To carry out \emph{identification experiments}, we used the first
image of each user as enrolment sample, and the remaining images
for evaluation. Given an evaluation sample, the user was recognized
by searching the enrolment samples of all the $K$ subjects in the
database for a match (one-to-many). As a result, the system returned
a ranked list of candidates.
For identification experiments, only users (eyes) with two or more
samples were considered. This resulted in $K$=162 available users,
and 764 different one-to-many trials (totalling 124,532
comparisons) per comparator and per scenario.

\subsection{Iris Recognition Algorithms}
\label{sec:matchers}

Iris recognition experiments were conducted using six different
algorithms according to the state of the art \cite{[Rathgeb16-USIT2]},
namely
%
1D log-Gabor filters (LG) \cite{[MasekThesis03]},
the SIFT operator (SIFT) \cite{[Lowe04]},
%
%
local intensity variations in iris textures (CR) \cite{[Rathgeb10]},
the Discrete-Cosine Transform (DCT) \cite{[Monro07]},
cumulative-sum-based grey change analysis (KO) \cite{[Ko07]}, and
Gabor spatial filters (QSW) \cite{[Ma03]}.

All algorithms (except SIFT) extract features from a normalized
rectangle image which is computed from the iris image using
the Daugman's rubber sheet model
\cite{[Daugman04]}. This normalization produces a 2D polar array of
20$\times$240  pixels, height$\times$width, (LG system) and 64$\times$512  pixels (others),
with horizontal dimensions of angular resolution, and vertical
dimensions of radial resolution. Feature encoding is implemented
according to the different feature extraction methods, leading to
fixed-length templates with are matched using distance measures
(details are given in the respective papers).
Rotation is accounted for by shifting the 2D polar array of
the query image in counter- and clock-wise direction and selecting
the lowest distance, which corresponds to the best match between the
two templates.
In the SIFT comparator, SIFT key points are directly extracted from the
iris image (without normalization), and the recognition metric is the
number of matched key points, normalized by the average number of
detected key-points in the two images under comparison.

We used open source code implementations of these algorithms.
The LG implementation is from the Libor Masek code
\cite{[MasekThesis03]}.
SIFT feature extraction and matching was carried out
using a free toolkit\footnote{http://vision.ucla.edu/$\sim$vedaldi/code/sift/assets/sift/index.html}
with the adaptations described in \cite{[Alonso09]} (particularly,
it includes a post-processing step to remove spurious matching
points using geometric constraints).
%
The remaining algorithms used are from the University of Salzburg
Iris Toolkit software package (USIT) \cite{[Rathgeb16-USIT2]}.

\begin{table*}
\scriptsize
\begin{center}
\begin{tabular}[t]{|c|c||c|c|c|c|c|c|c||c|c|c|c|c|c|c||}


\multicolumn{16}{c}{} \\ \cline{3-9} \cline{10-16}

\multicolumn{2}{c||}{} & \multicolumn{7}{c||}{Full image} & \multicolumn{7}{|c||}{Unwrapped iris region}\\
\cline{3-9} \cline{10-16}

\multicolumn{16}{c}{} \\ \cline{1-1} \cline{5-8} \cline{12-15}

\textbf{LR size} & \multicolumn{3}{c}{} &
\multicolumn{4}{|c|}{\textbf{Our method} (patch size)} &
\multicolumn{1}{c}{}&
\multicolumn{2}{c}{}
& \multicolumn{4}{|c|}{\textbf{Our method} (patch size)} &
\multicolumn{1}{c}{}\\
\cline{3-9} \cline{10-16}

(scaling) & & \textbf{bilinear} & \textbf{bicubic} & 1/4 & 1/8 &
1/16
& 1/32 & \textbf{diff} & bilinear & bicubic & 1/4 & 1/8 & 1/16 & 1/32 & \textbf{diff} \\
\hline \hline

115$\times$115 & psnr & 33 & 34.04 & 34.23 & \textbf{34.65} & 34.62 & 34.11 & +0.61 &  36.94  & 38.22 & 38.77 & \textbf{39.15} & 39.1 & 38.59  & +0.93 \\
 \cline{2-16}

(1/2) & ssim & 0.91 & 0.93 & 0.92 & 0.93 & \textbf{0.94} & 0.93 & +0.01  & 0.96 & \textbf{0.97} & \textbf{0.97} & \textbf{0.97} & \textbf{0.97} & \textbf{0.97}   & +0.00 \\
\hline \hline

57$\times$57  & psnr & 28.36 & 29.18 & \textbf{29.91} & 29.9 & 29.53 & 28.78 & +0.73 & 31.64  & 32.35 & 32.69 & \textbf{32.72} & 32.38 & 31.83  & +0.37 \\
 \cline{2-16}

(1/4) & ssim & 0.79 & 0.8 & 0.8 & \textbf{0.81} & 0.8 & 0.78 & +0.01  & 0.85  & 0.87 & \textbf{0.88} & \textbf{0.88} & \textbf{0.88} & 0.86  & +0.01  \\
\hline \hline

39$\times$39  & psnr & 26.21 & 26.8 & 27.98 & \textbf{28.05} & 27.86 & 27.61 & +1.25  & 29.61  & 30.21 & 30.71 & \textbf{30.85}  & 30.73  & 30.64  & +0.64 \\
 \cline{2-16}

(1/6) & ssim & 0.73 & 0.74 & 0.74 & \textbf{0.75} & \textbf{0.75} & \textbf{0.75} & +0.01  & 0.79  & 0.81 & 0.81 & \textbf{0.82} & \textbf{0.82}  & \textbf{0.82}  & +0.01  \\
\hline \hline

29$\times$29  & psnr & 24.86 & 25.33 & \textbf{26.73} & 26.55 & 26.16 & - & +1.4  & 28.18  & 28.74 & \textbf{29.57} & 29.49 & 29.2 & -  & +0.83 \\
 \cline{2-16}

(1/8) & ssim & 0.69 & 0.7 & \textbf{0.71} & \textbf{0.71} & 0.7 & - & +0.01 & 0.74 & 0.75 & \textbf{0.77} & \textbf{0.77} & 0.76 & -  & +0.02  \\
\hline \hline

23$\times$23  & psnr & 23.94 & 24.41 & \textbf{25.76} & 25.62 & 25.24 & - & +1.35 & 27.09 & 27.63 & \textbf{28.69} & 28.47 & 28.11 & -  &  +1.06  \\
 \cline{2-16}

(1/10) & ssim & 0.67 & 0.68 & \textbf{0.69} & \textbf{0.69} & 0.67 & - & +0.01  & 0.7 & 0.71 & \textbf{0.73} & 0.72 & 0.71 & -   & +0.02 \\
\hline \hline

19$\times$19  & psnr & 23.24 & 23.71 & \textbf{25.12} & 24.95 & 24.57 & - & +1.41 & 26.21 & 26.71 & \textbf{27.88} & 27.53 & 27.18 & -  & +1.17 \\
 \cline{2-16}

(1/12) & ssim & 0.65 & 0.66 & \textbf{0.68} & 0.67 & 0.65 & - & +0.02 & 0.66 & 0.68 & \textbf{0.69} & 0.68 & 0.67 & -  & +0.01 \\
\hline \hline

17$\times$17  & psnr & 22.85 & 23.32 & \textbf{24.69} & 24.05 & - & - & +1.37 & 25.72 & 26.22 & \textbf{27.39} & 26.53 & - & -  & +1.17 \\
 \cline{2-16}

(1/14) & ssim & 0.64 & 0.65 & \textbf{0.66} & 0.64 & - & - & +0.01 & 0.65 & 0.66 & \textbf{0.67} & 0.63 & - & -  & +0.01  \\
\hline \hline

15$\times$15  & psnr & 22.39 & 22.86 & \textbf{24.32} & 24.17 & - & - & +1.46  & 25.16 & 25.64 & \textbf{26.99} & 26.79 & - & -  & +1.35 \\
 \cline{2-16}

(1/16) & ssim & 0.64 & 0.64 & \textbf{0.66} & 0.65 & - & - & +0.02 & 0.63   & 0.64 & \textbf{0.66} & 0.65 & - & -  & +0.02  \\
\hline \hline



\end{tabular}
\caption{Hallucination results (PSNR and SSIM values) with
different downsampling factors and patch sizes (average
values on the test dataset). Patch size is indicated in proportion
to the size of the iris image. The best results for each downsampling
factor are marked in bold. `Diff' refers to the difference between the best PCA case and the bicubic method.} \label{tabla:results-psnr-ssim}
\end{center}
\end{table*}
\normalsize

\begin{figure*}[t]
     \centering
     \includegraphics[width=.95\textwidth]{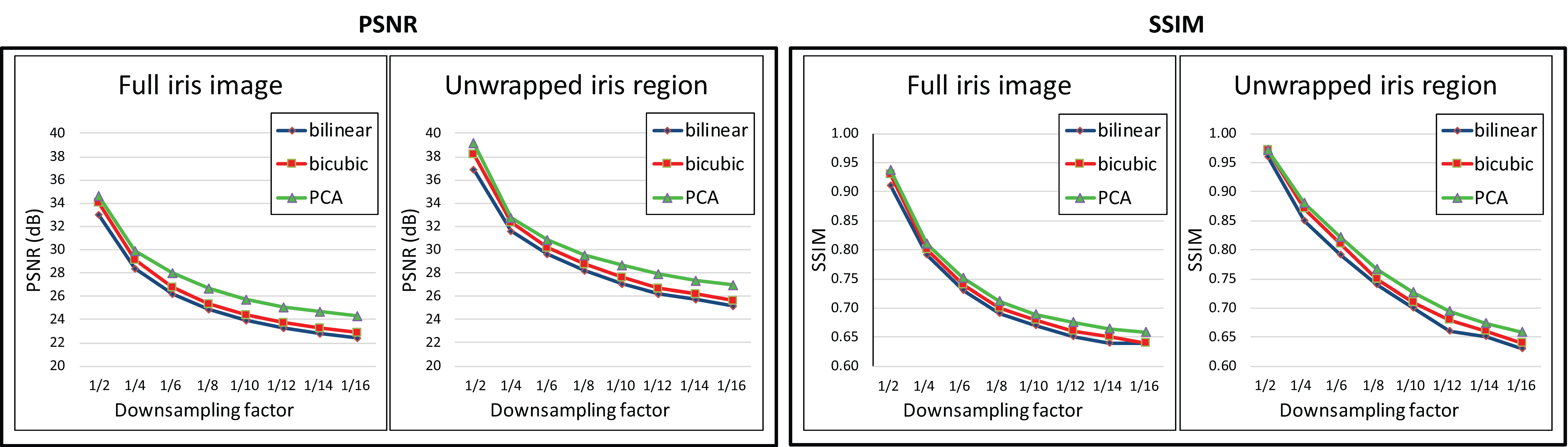}
     \caption{Hallucination results with different downsampling factors (left: PSNR values, right: SSIM values). The best PCA case for each downsampling factor is shown (marked in bold in Table~\ref{tabla:results-psnr-ssim}).}
     \label{fig:results-psnr-ssim}
\end{figure*}

\section{Experimental Results}
\label{sect:results}

\subsection{Image Fidelity}
\label{sect:results-fidelity}

This section reports
the performance of the hallucination algorithms by
measuring the Peak Signal-to-Noise ratio (PSNR) and the Structural Similarity (SSIM) index
%
%
between the hallucinated image
and its corresponding high-resolution reference image of the test set.
We used MATLAB's \texttt{psnr} and \texttt{ssim} functions for this purpose.
These are the metrics usually employed in the super-resolution literature \cite{[Nasrollahi2014]}.
The PSNR is a measure of the ratio (in dBs) between the maximum possible power of a signal
and the power of corrupting noise that affects the fidelity of its representation.
The noise in this case is defined as the difference between the reference high-resolution image $\overline{Y}$,
and its estimation $\tilde{Y}$ by the reconstruction algorithm.
A higher PSNR generally indicates that the reconstruction is of higher quality.
If the two images are identical, $\overline{Y}=\tilde{{Y}}$, then $\textrm{PSNR} (\overline{Y},\tilde{{Y}})=\infty$.
%
%
The SSIM index, on the other hand, is a perception-based model that considers image degradation as a perceived change in structural information (luminance and contrast).
This is achieved by using first and second order statistics of grey values on local image windows.
The SSIM index is a decimal value between -1 and 1, and value 1 is only reachable in the case of two identical images.

Table~\ref{tabla:results-psnr-ssim} reports the average PSNR and SSIM values of all images in the test set.
We report PSNR and SSIM metrics in two cases: $i$) between the full reconstructed image and its reference high-resolution image; and $ii$) between the normalized polar image versions of these two images (with size 20$\times$240 pixels), computed according to the
%
Daugman's rubber sheet model \cite{[Daugman04]}.
%
%
Examples of hallucinated iris
images can also be seen in Figure~\ref{fig:images-example}
(only for a selection of downsampling factors for the sake of space).
In the described PCA method, the size of 
local image patches is an important parameter.
In order to test the influence of this parameter,
we evaluated the performance of the PCA algorithm
with different patch sizes.
%
In particular, we tested patches of size equal to
1/4, 1/8, 1/16 and 1/32 of the
iris image.
We defined the patch size in proportion to the
dimensions of the 
iris image to ensure that they cover the same
relative size across different scaling factors.
Overlapping between
patches was set to 1/3. 
%

\begin{table*}
\scriptsize
\begin{center}
\begin{tabular}[htb]{|c|c|c|c|c|c|c|c|c|c|c|c|c|}

\multicolumn{4}{c}{} & \multicolumn{4}{c}{Scenario 1 (original vs.
down-sampled)} & \multicolumn{1}{c}{} & \multicolumn{4}{c}{Scenario
2 (down-sampled vs. down-sampled)} \\ \cline{5-8} \cline{10-13}

\multicolumn{4}{c}{} & \multicolumn{4}{c}{Downsampling} &
\multicolumn{1}{c}{} &
\multicolumn{4}{c}{Downsampling} \\
\cline{1-1} \cline{3-3} \cline{5-8} \cline{10-13}

 & & \textbf{} & & 115$\times$115  &
57$\times$57 & 29$\times$29 & 15$\times$15
& & 115$\times$115  &
57$\times$57 & 29$\times$29 & 15$\times$15 \\

\textbf{Comparator} & & \textbf{Method} & & (1/2) & (1/4) & (1/8) &
 (1/16)
& & (1/2) & (1/4) & (1/8) &
 (1/16) \\

\hline \hline

 & & Bilinear & & \textbf{0.69\%} & 0.69\% & 1.61\% & 10.39\% &  & \textbf{0.61\%} & 0.76\% & 2.38\% & 11.03\% \\

LG & & Bicubic & & \textbf{0.69\%} & \textbf{0.68\%} & 1.42\% & 9.59\% &  & 0.73\% & \textbf{0.65\%} & 1.88\% & 11.25\% \\ \cline{3-3} \cline{5-8} \cline{10-13}

 & & Our & & 0.76\% & 0.8\% & \textbf{1.11\%} & \textbf{7.29\%} &  & 0.73\% & 0.69\% & \textbf{1.18\%} & \textbf{4.79\%} \\

  & & (var) & & (+10.1\%) & (+17.6\%) & \textbf{(-21.8\%)} & \textbf{(-24\%)} & & (+19.7\%)& (+6.2\%)& \textbf{(-37.2\%)}& \textbf{(-56.6\%)} \\

\hline \hline

 & & Bilinear & & 4.05\% & 10.42\% & 28.23\% & 50.52\% &  & \textbf{3.01\%} & 4.26\% & 14.82\% & 41.66\% \\

SIFT & & Bicubic & & \textbf{3.51\%} & 7.41\% & 24.99\% & 47.33\% &  & 3.13\% & \textbf{3.08\%} & 11.6\% & 36.37\% \\ \cline{3-3} \cline{5-8} \cline{10-13}

 & & Our & & 4.17\% & \textbf{4.74\%} & \textbf{15.65\%} & \textbf{36.67\%} &  & 3.9\% & 3.11\% & \textbf{7.46\%} & \textbf{18.98\%} \\

  & & (var) & & (+18.8\%) & \textbf{(-36\%)} & \textbf{(-37.4\%)} & \textbf{(-22.5\%)} & & (+29.6\%)& (+1\%)& \textbf{(-35.7\%)}& \textbf{(-47.8\%)} \\
\hline \hline

%
%

 & & Bilinear & & 10.12\% & 13.06\% & 20.96\% & 33.74\% &  & 8.86\% & 9.24\% & 13\% & 15.76\% \\

CR & & Bicubic & & \textbf{9.85\%} & 11.98\% & 19.45\% & 34.1\% &  & \textbf{8.84\%} & \textbf{7.83\%} & 10.98\% & 17.49\% \\ \cline{3-3} \cline{5-8} \cline{10-13}

 & & Our & & 10.41\% & \textbf{11.33\%} & \textbf{16.88\%} & \textbf{28.1\%} &  & 10.1\% & 9.04\% & \textbf{9.98\%} & \textbf{14.44\%} \\

  & & (var) & & (+5.7\%) & \textbf{(-5.4\%)} & \textbf{(-13.2\%)} & \textbf{(-16.7\%)} & & (+14.3\%)& (+15.5\%)& \textbf{(-9.1\%)}& \textbf{(-8.4\%)} \\
\hline \hline

 & & Bilinear & & \textbf{2.15\%} & 3.59\% & 16.64\% & 44.51\% &  & \textbf{1.77\%} & 2.55\% & 9.25\% & 17.77\% \\

DCT & & Bicubic & & 2.16\% & 2.95\% & 14.05\% & 44.03\% &  & 2.06\% & 2.14\% & 7.63\% & 18.11\% \\ \cline{3-3} \cline{5-8} \cline{10-13}

 & & Our & & 2.19\% & \textbf{2.73\%} & \textbf{10.39\%} & \textbf{39.09\%} &  & 2.11\% & \textbf{2.13\%} & \textbf{6.75\%} & \textbf{11.72\%} \\

  & & (var) & & (+1.9\%) & \textbf{(-7.5\%)} & \textbf{(-26\%)} & \textbf{(-11.2\%)} & & (+19.2\%)& \textbf{(-0.5\%)}& \textbf{(-11.5\%)}& \textbf{(-34\%)} \\
\hline \hline

 & & Bilinear & & \textbf{12.18\%} & 12.68\% & 14.39\% & 18.41\% &  & 12.48\% & 12.92\% & 14.08\% & 15\% \\

KO & & Bicubic & & 12.55\% & 12.74\% & 13.95\% & 17.49\% &  & \textbf{12.37\%} & 12.49\% & 13.75\% & 14.92\% \\ \cline{3-3} \cline{5-8} \cline{10-13}

 & & Our & & 12.49\% & \textbf{12.45\%} & \textbf{13.44\%} & \textbf{16.63\%} &  & 12.54\% & \textbf{12.29\%} & \textbf{12.8\%} & \textbf{13.41\%} \\

  & & (var) & & (+2.5\%) & \textbf{(-1.8\%)} & \textbf{(-3.7\%)} & \textbf{(-4.9\%)} & & (+1.4\%)& \textbf{(-1.6\%)}& \textbf{(-6.9\%)}& \textbf{(-10.1\%)} \\
\hline \hline

 & & Bilinear & & \textbf{0.7\%} & 0.76\% & 2.09\% & 14.47\% &  & \textbf{0.76\%} & 0.84\% & 3.06\% & 12.49\% \\

QSW & & Bicubic & & 0.74\% & 0.77\% & 1.8\% & \textbf{12.13\%} &  & 0.8\% & 0.8\% & 2.93\% & 12.95\% \\ \cline{3-3} \cline{5-8} \cline{10-13}

 & & Our & & 0.76\% & \textbf{0.75\%} & \textbf{1.67\%} & 12.31\% &  & \textbf{0.76\%} & \textbf{0.77\%} & \textbf{1.91\%} & \textbf{5.55\%} \\

  & & (var) & & (+8.6\%) & \textbf{(-1.3\%)} & \textbf{(-7.2\%)} & (+1.5\%) & & (0\%)& \textbf{(-3.8\%)}& \textbf{(-34.8\%)}& \textbf{(-55.6\%)} \\
\hline \hline

\end{tabular}
\caption{Verification results (EER values) of the two scenarios
considered for different downsampling factors. The best case for each comparator and for each downsampling factor is marked in bold. The relative EER variation of PCA with respect to bilinear/bicubic (best of the two) is also given.}
\label{tabla:results-EER}
\end{center}
\end{table*}
\normalsize

\begin{figure*}[htb]
     \centering
     \includegraphics[width=.98\textwidth]{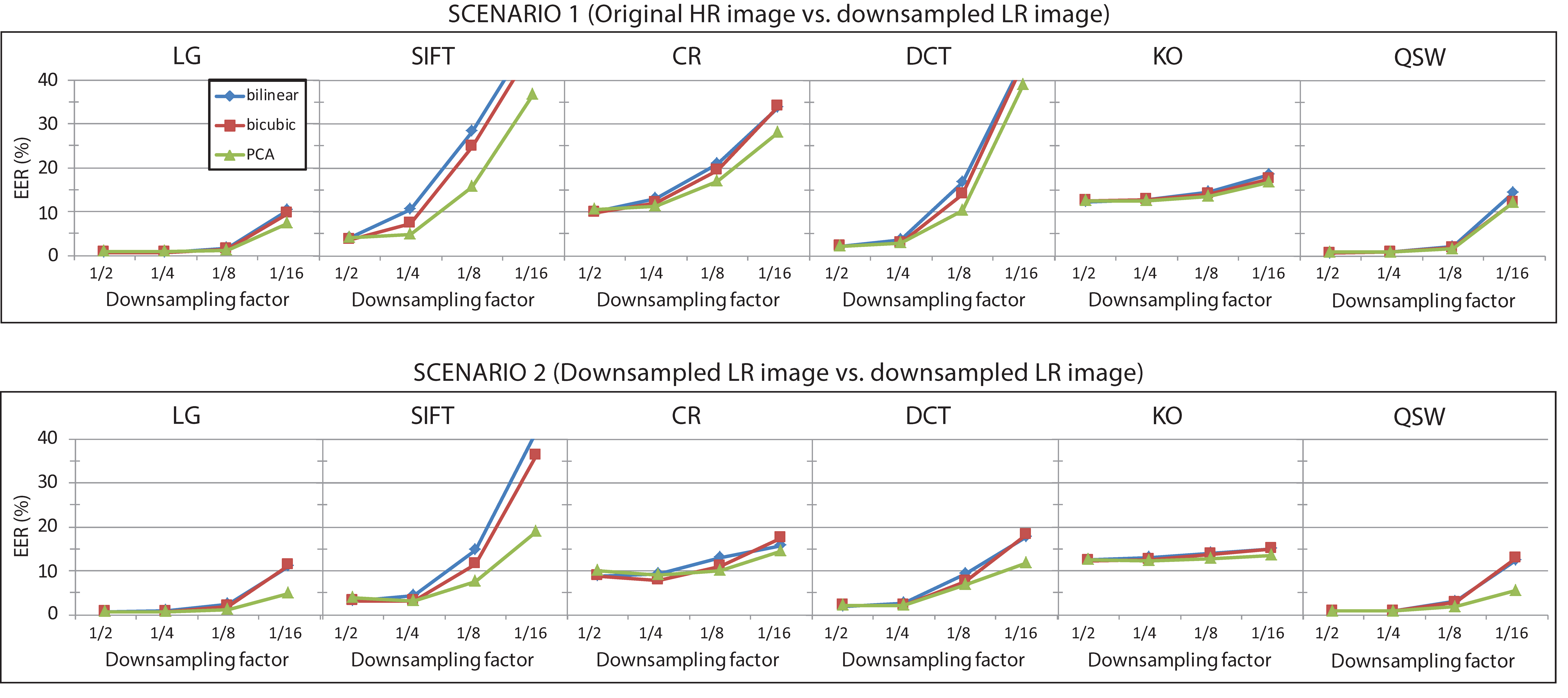}
     \caption{Verification results (EER values) with different downsampling factors.}
     \label{fig:results-eer}
\end{figure*}

Table~\ref{tabla:results-psnr-ssim} shows in bold the best result for each downsampling factor.
As it can be seen, 
the eigen-patch hallucination method outperforms the bilinear and
bicubic interpolations at all resolutions.
Also, among bilinear and bicubic interpolations, the latter gives the best results.
We can also observe that, as resolution decreases, the gain of the PCA method becomes larger, at least as measured by the PSNR.
%
This can be
better assessed in Figure~\ref{fig:results-psnr-ssim}, where we plot
the results of Table~\ref{tabla:results-psnr-ssim} (the best
PCA case for each downsampling factor is selected).
At a resolution of 115$\times$115 pixels, the PSNR gain of PCA is of 0.61dB over bicubic (full image) and 0.93dB (iris region), and it becomes larger at lower resolutions. At a resolution of 15$\times$15 pixels, the gain reaches 1.46dB (full image) and 1.35dB (iris region), showing the advantage of the utilized eigen-patch hallucination method at very low resolutions.
%

It is also worth noting that, although the average SSIM values are smaller as resolution decreases, the gain of PCA over bicubic remains between 0.01 and 0.02.
In Figure~\ref{fig:images-example}, it can be observed that images reconstructed with bilinear or bicubic methods
have much more blur than those reconstructed with PCA when the resolution decreases.
While this is reflected in a higher gain in PSNR, it is no the case with the SSIM.
As it has been pointed out in previous studies,
image quality metrics like PSNR or SSIM do not have the same sensitivity to image degradations \cite{[AlNajjar12],[Hore10]}.
Although iris images reconstructed with PCA have better subjective quality, this is not captured well by the SSIM index.
%
%


Regarding the appropriate patch size, it can be observed in
Table~\ref{tabla:results-psnr-ssim} that the best results are
obtained with a bigger patch. In particular, 1/4 is the best
size at very low-resolutions.
Using a bigger patch results in less artefacts due to overlapping
patches being stitched together. This can be seen for example in
Figure~\ref{fig:images-example} (image of size 57$\times$57 pixels),
where more artefacts appear for patch sizes of 1/16 or 1/32. It
also has computational implications, because there are
less patches to process per image, although they have a bigger size.

Given than a patch size of 1/4 consistently produces top results
in the experiments of this section, we will employ this patch size with PCA
in the remainder of this paper.
This patch size has also been observed to produce top results in
previously published verification experiments using the LG comparator \cite{[Alonso15b]}.


\begin{figure*}[htb]
     \centering
     \includegraphics[width=.98\textwidth]{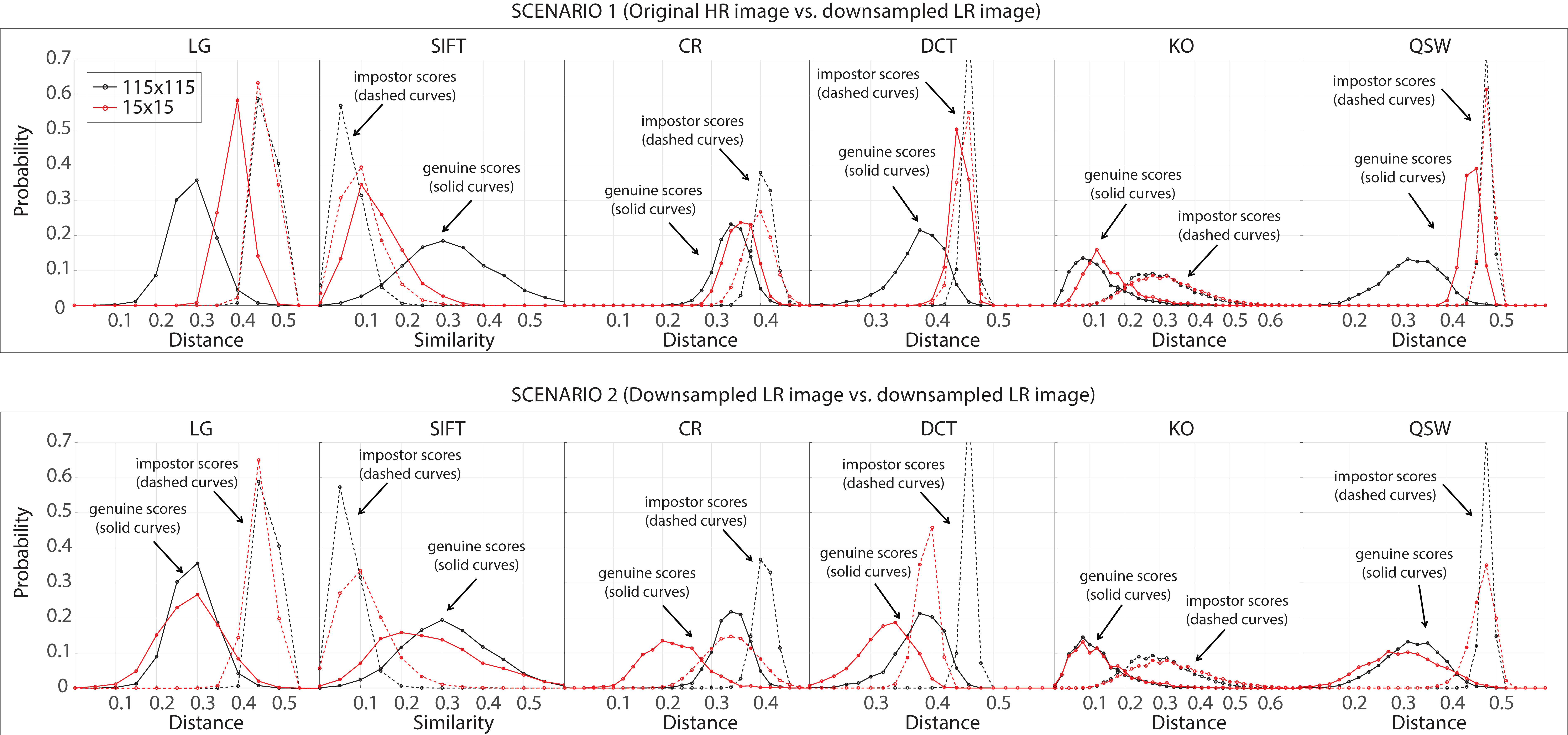}
     \caption{
     Score distributions with PCA enhancement for a downsampling factor of 1/2 (image size 115$\times$115 pixels, black curves) and 1/16 (image size 15$\times$15 pixels, red curves).}
     \label{fig:results-score-distribution}
\end{figure*}

\begin{figure*}[htb]
     \centering
     \includegraphics[width=.98\textwidth]{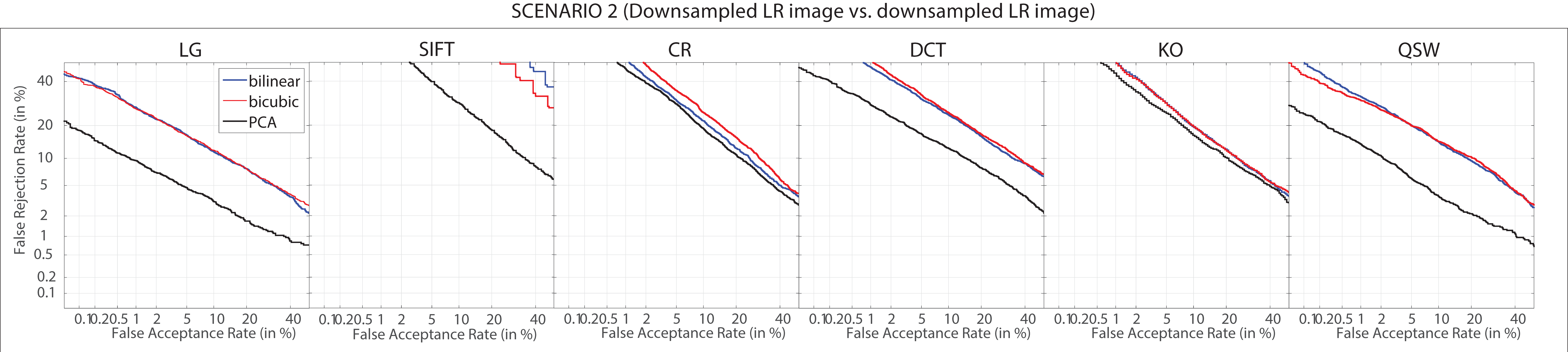}
     \caption{Verification results (DET curves) of the iris comparators.
       Results are given for scenario 2 and a downsampling factor of 1/16 (image size 15$\times$15 pixels ).}
     \label{fig:results-score-det}
\end{figure*}

\subsection{Biometric Verification}

Next, we report results of the verification experiments using hallucinated iris images.
Verification results of the proposed PCA method and the bilinear/bicubic interpolations
are given in Table~\ref{tabla:results-EER} and Figure~\ref{fig:results-eer}.
We adopt the Equal Error Rate (EER) as the measure of accuracy.
Due to space constraints, we only report a selection of downsampling factors
(those shown in Figure~\ref{fig:images-example}).
%
%
%
In Table~\ref{tabla:results-EER}, we also mark in bold
the best EER value for each comparator and for each downsampling factor.
In addition, the relative EER variation of PCA with respect to bilinear/bicubic
(best of the two) is given in brackets.

\subsubsection{Interpolation Method Analysis}

It can be seen that the PCA method results in better verification performance than bilinear/bicubic
interpolations when the resolution decreases.
This highlights the benefits of using trained methods to enhance very low-resolution iris images.
At an image size of 29$\times$29 pixels (downsampling of 1/8),
any given comparator produces better performance with PCA than with bilinear or bicubic interpolations.
In scenario 1, for example,
the EER reduction of PCA is between 3.7\% (KO comparator) and 37.4\% (SIFT); whereas in scenario 2, it is between 6.9\% (KO) and 37.2\% (LG).
For some comparators, the improvement with PCA is evident even at a resolution of 57$\times$57 pixels (downsampling of 1/4). For example, the EER reduction in this case is of 36\% with the SIFT comparator, and of 7.5\% with the DCT system (both in scenario 1).
%
%
      But the biggest benefits of PCA occur at very low-resolutions
      (15$\times$15 pixels, or downsampling of 1/16). 
      Here, EER improvements are significantly higher, as reflected by the wider gap
      between curves in Figure~\ref{fig:results-eer}.
      %
      %
      With such small image size, an impressive EER reduction of 56-57\%
      is achieved with LG and QSW,
      followed by 47.8\% with SIFT, and 34\% with DCT (all in scenario 2).
      %

\subsubsection{Scenario Analysis}
\label{sect:scenario-analysis-verif}

In this sub-section, we study the difference between the two scenarios of operation analysed.
It can be observed in Figure~\ref{fig:results-eer} that
  the performance of scenario 1 and 2 is approximately equal up to a certain resolution.
    Then, if resolution is further reduced,
    scenario 2 performs much better than scenario 1.
    This is observed with all comparators, being the only difference the cut-off point.
    For example, with LG, KO, or QSW, the difference between both scenarios only becomes substantial
    at a factor of 1/16; but with DCT the difference starts at 1/8, and even earlier with SIFT and CR.

    We further analyse this effect by plotting
    in Figure~\ref{fig:results-score-distribution} the score distributions with PCA of the following
    two cases:
    moderate downsampling (factor 1/2, image size 115$\times$115 pixels, black curves),
    and extreme downsampling (factor 1/16, image size 15$\times$15 pixels, red curves).
    %
    Note that most comparators employ a distance measure, meaning that genuine distributions
    are on the left side of the plot,
    but one comparator (SIFT) employ a similarity measure, so its genuine distributions
    are on the right side.

    In scenario 1, we observe that in most comparators, the distribution of genuine scores is shifted
    towards the impostor distribution as resolution decreases.
    This suggests that the
    PCA reconstruction algorithm is not able to fully recover
    the information found in the original high
    resolution image, at least measured by the features employed by our comparators.
    As a result,
    the similarity between high-resolution and reconstructed images of the same user is reduced
    (recall that in scenario 1, enrolment samples were taken from high-resolution images,
    whereas query samples were taken from reconstructed images).
    This is consistent with the results of Figure~\ref{fig:results-psnr-ssim}, which show that the PSNR and SSIM values between a high-resolution image and its reconstructed counterpart decreases with the resolution.
    On the other hand, impostor distributions lie in a similar range in most cases,
    regardless of the resolution. 
    %

    In scenario 2, on the contrary, the distribution of genuine scores
    is not shifted significantly towards the impostor distribution
    as the resolution changes. Instead, the relative difference between
    the two distributions is maintained.
    In this scenario,
    enrolment and query images undergo the same
    downsampling and upsampling procedure,
    regardless whether they come from genuine or impostor trials.
    Therefore, it has sense that the relative difference between
    genuine users and impostors is maintained.
    However, a collateral effect, given by the loss of information
    when images are down-sampled and reconstructed, is that genuine and score distributions
    are more spread.
    This explains the worse EER performance  at low resolution.
    An exception to these observations is SIFT.
    With this comparator, the genuine and impostor distributions
    become significantly closer to each other in scenario 2.
    For this reason, this is the comparator
    whose performance is most significantly degraded
    with the resolution (see Figure~\ref{fig:results-eer}).
    %
    %
    Another exception, but in the opposite direction, is the KO comparator.
    This comparator shows a high resilience
    to changes in resolution, as it can be seen by its
    nearly `flat' behaviour in Figure~\ref{fig:results-eer}.
    As a result, its score distributions are in the same range.
    It should be remarked, however, that its EER at high-resolution
    is already above 12\%, while other comparators start below 1\%.
    %

\subsubsection{Comparator Analysis}

We further look into the differences between the individual comparators considered in this paper.
As it can be seen in Figure~\ref{fig:results-eer},
  the performance of any comparator is not degraded significantly
  until a downsampling factor of at least 1/8 (image size of 29$\times$29 pixels).
  This implies that the size of both gallery and probe images could be kept
  low without sacrificing performance, and then simply
  up-sampled with the bicubic or even bilinear method.
  This has positive implications when 
  lower storage
  or data transmission capabilities are required. 
  %
  Some comparators (LG, QSW) even show a praiseworthy performance with
  a downsampling factor of 1/8, having in this case an EER of 1.11\%/1.67\% in scenario 1,
  and 1.18\%/1.91\% in scenario 2.
  It should be noted as well that these two comparators are based on Gabor wavelets,
  so it is reasonable that they behave in a similar fashion.
  It is also remarkable the impressive low EER figures obtained
  with these comparators when considering very low-resolution images
  with a size of only 15$\times$15 pixels.
  For example, an EER of 4.78\% is obtained with the LG comparator,
  and 5.55\% with QSW in scenario 2.
  %
    %
    Also, as mentioned above, the KO comparator shows a high resilience
    to changes in resolution, with the EER degrading only from 12.54\% to 13.41\%
    in scenario 2.

  In terms of absolute performance, the best comparator at any resolution is LG,
  followed closely by QSW (recall that both are based on Gabor wavelets).
  The third one is DCT, followed by SIFT, although performance of these two
  degrades significantly at very low-resolutions.
  For example, DCT has an EER of $\sim$2\% at a resolution of 1/2, but it goes up to 39.09\% at a resolution of 1/16 in scenario 1, and to 11.72\% in scenario 2.
  %
  CR and KO are the worst performing comparators. They both start with an EER of 10\% or higher already at a resolution of 1/2,
  although they show higher resiliency to reductions in resolution, as seen earlier.

For completeness, we provide in Figure~\ref{fig:results-score-det} the DET curves of all comparators
(only scenario 2 and a downsampling factor of 1/16). Here, it can be observed the gain of the
trained PCA enhancement in comparison to bilinear and bicubic interpolations. A consistent
improvement across all ranges of the DET curve is obtained with most comparators.


\begin{table*}
\scriptsize
\begin{center}
\begin{tabular}[htb]{|c|c|c|c|c|c|c|c|c|c|c|c|c|}

\multicolumn{4}{c}{} & \multicolumn{4}{c}{Scenario 1 (original vs.
down-sampled)} & \multicolumn{1}{c}{} & \multicolumn{4}{c}{Scenario
2 (down-sampled vs. down-sampled)} \\ \cline{5-8} \cline{10-13}

\multicolumn{4}{c}{} & \multicolumn{4}{c}{Downsampling} &
\multicolumn{1}{c}{} &
\multicolumn{4}{c}{Downsampling} \\
\cline{1-1} \cline{3-3} \cline{5-8} \cline{10-13}

 & & \textbf{} & & 115$\times$115  &
57$\times$57 & 29$\times$29 & 15$\times$15 & & 115$\times$115  &
57$\times$57 & 29$\times$29 & 15$\times$15 \\

\textbf{Comparator} & & \textbf{Method} & & (1/2) & (1/4) & (1/8) &
 (1/16)
& & (1/2) & (1/4) & (1/8) &
 (1/16) \\

\cline{1-1} \cline{3-3} \cline{5-8} \cline{10-13}

 & & Bilinear & & \textbf{98.82\%} & \textbf{98.95\%} & 95.94\% & 54.58\% &  & \textbf{99.21\%} & 99.08\% & 92.8\% & 66.23\% \\

LG & & Bicubic & & \textbf{98.82\%} & \textbf{98.95\%} & 96.34\% & 59.82\% &  & 99.08\% & \textbf{99.21\%} & 94.5\% & 65.45\% \\ \cline{3-3} \cline{5-8} \cline{10-13}

 & & Our & & 98.69\% & \textbf{98.95\%} & \textbf{97.51\%} & \textbf{73.43\%} &  & 98.82\% & \textbf{99.21\%} & \textbf{98.04\%} & \textbf{84.16\%} \\

   & & (var) & & (-0.1\%) & (0\%)& \textbf{(+1.2\%)}& \textbf{(+22.8\%)} & & (-0.4\%)& (0\%)& \textbf{(+3.7\%)}& \textbf{(+27.1\%)} \\

 \hline \hline

 & & Bilinear & & 88.74\% & 71.73\% & 13.48\% & 0.92\% &  & 91.88\% & 87.17\% & 45.94\% & 5.5\% \\

SIFT & & Bicubic & & \textbf{89.53\%} & 79.06\% & 22.91\% & 0.65\% &  & \textbf{93.19\%} & \textbf{92.8\%} & 60.34\% & 9.95\% \\\cline{3-3} \cline{5-8} \cline{10-13}

 & & Our & & 88.09\% & \textbf{86.91\%} & \textbf{51.44\%} & \textbf{5.37\%} &  & 88.74\% & 92.54\% & \textbf{78.01\%} & \textbf{40.71\%} \\

   & & (var) & & (-1.6\%) & \textbf{(+9.9\%)}& \textbf{(+124.5\%)}& \textbf{(+483.7\%)} & & (-4.8\%)& (-0.3\%)& \textbf{(+29.3\%)}& \textbf{(+309.1\%)} \\

\hline \hline

%
%

 & & Bilinear & & 77.23\% & 61.78\% & 28.27\% & 4.58\% &  & \textbf{84.82\%} & 80.63\% & 64.92\% & 46.34\% \\

CR & & Bicubic & & \textbf{79.71\%} & 68.19\% & 34.29\% & 4.32\% &  & 84.16\% & \textbf{85.34\%} & 69.24\% & 43.59\% \\\cline{3-3} \cline{5-8} \cline{10-13}

 & & Our & & 78.93\% & \textbf{77.36\%} & \textbf{46.6\%} & \textbf{14.53\%} &  & 80.5\% & 83.38\% & \textbf{70.03\%} & \textbf{53.27\%} \\

   & & (var) & & (-1\%) & \textbf{(+13.4\%)}& \textbf{(+35.9\%)}& \textbf{(+217.2\%)} & & (-5.1\%)& (-2.3\%)& \textbf{(+1.1\%)}& \textbf{(+15\%)} \\

\hline \hline

 & & Bilinear & & 96.73\% & 90.97\% & 26.18\% & 0.39\% &  & \textbf{97.77\%} & 96.34\% & 80.5\% & 52.62\% \\

DCT & & Bicubic & & 96.99\% & 92.93\% & 37.04\% & 0.39\% &  & 96.99\% & \textbf{96.99\%} & 83.12\% & 54.32\% \\\cline{3-3} \cline{5-8} \cline{10-13}

 & & Our & & \textbf{97.12\%} & \textbf{94.63\%} & \textbf{61.52\%} & \textbf{0.92\%} &  & 97.12\% & 95.55\% & \textbf{83.38\%} & \textbf{71.07\%} \\

   & & (var) & & \textbf{(+0.1\%)} & \textbf{(+1.8\%)}& \textbf{(+66.1\%)}& \textbf{(+135.9\%)} & & (-0.7\%)& (-1.5\%)& \textbf{(+0.3\%)}& \textbf{(+30.8\%)} \\

\hline \hline

 & & Bilinear & & \textbf{56.41\%} & 54.45\% & 46.99\% & 27.09\% &  & 55.76\% & \textbf{55.76\%} & 46.6\% & 37.83\% \\

KO & & Bicubic & & 55.89\% & 54.97\% & 48.17\% & 30.37\% &  & \textbf{56.15\%} & 54.58\% & 49.21\% & 40.84\% \\\cline{3-3} \cline{5-8} \cline{10-13}

 & & Our & & 55.5\% & \textbf{56.41\%} & \textbf{50.92\%} & \textbf{37.17\%} &  & 54.71\% & 55.63\% & \textbf{52.49\%} & \textbf{48.56\%} \\

   & & (var) & & (-1.6\%) & \textbf{(+2.6\%)}& \textbf{(+5.7\%)}& \textbf{(+22.4\%)} & & (-2.6\%)& (-0.2\%)& \textbf{(+6.7\%)}& \textbf{(+18.9\%)} \\

\hline \hline

 & & Bilinear & & \textbf{99.21\%} & 99.08\% & 95.16\% & 32.59\% &  & 99.21\% & 99.08\% & 90.45\% & 60.73\% \\

QSW & & Bicubic & & 99.08\% & \textbf{99.35\%} & 95.55\% & 37.7\% &  & 99.21\% & \textbf{99.35\%} & 92.28\% & 60.6\% \\\cline{3-3} \cline{5-8} \cline{10-13}

 & & Our & & 99.08\% & 98.95\% & \textbf{97.38\%} & \textbf{42.8\%} &  & \textbf{99.35\%} & 99.08\% & \textbf{95.29\%} & \textbf{77.23\%} \\

   & & (var) & & (-0.1\%) & (-0.4\%)& \textbf{(+1.9\%)}& \textbf{(+13.5\%)} & & \textbf{(+0.1\%)}& (-0.3\%)& \textbf{(+3.3\%)}& \textbf{(+27.2\%)} \\

\hline \hline

\end{tabular}
\caption{Identification results (Top-1 accuracy) of the two
scenarios considered for different downsampling factors. The best case for each comparator and for each downsampling factor is marked in bold. The relative Top-1 variation of PCA with respect to bilinear/bicubic (best of the two) is also given.}
\label{tabla:results-rank1}
\end{center}
\end{table*}
\normalsize

\begin{figure*}[htb]
     \centering
     \includegraphics[width=.98\textwidth]{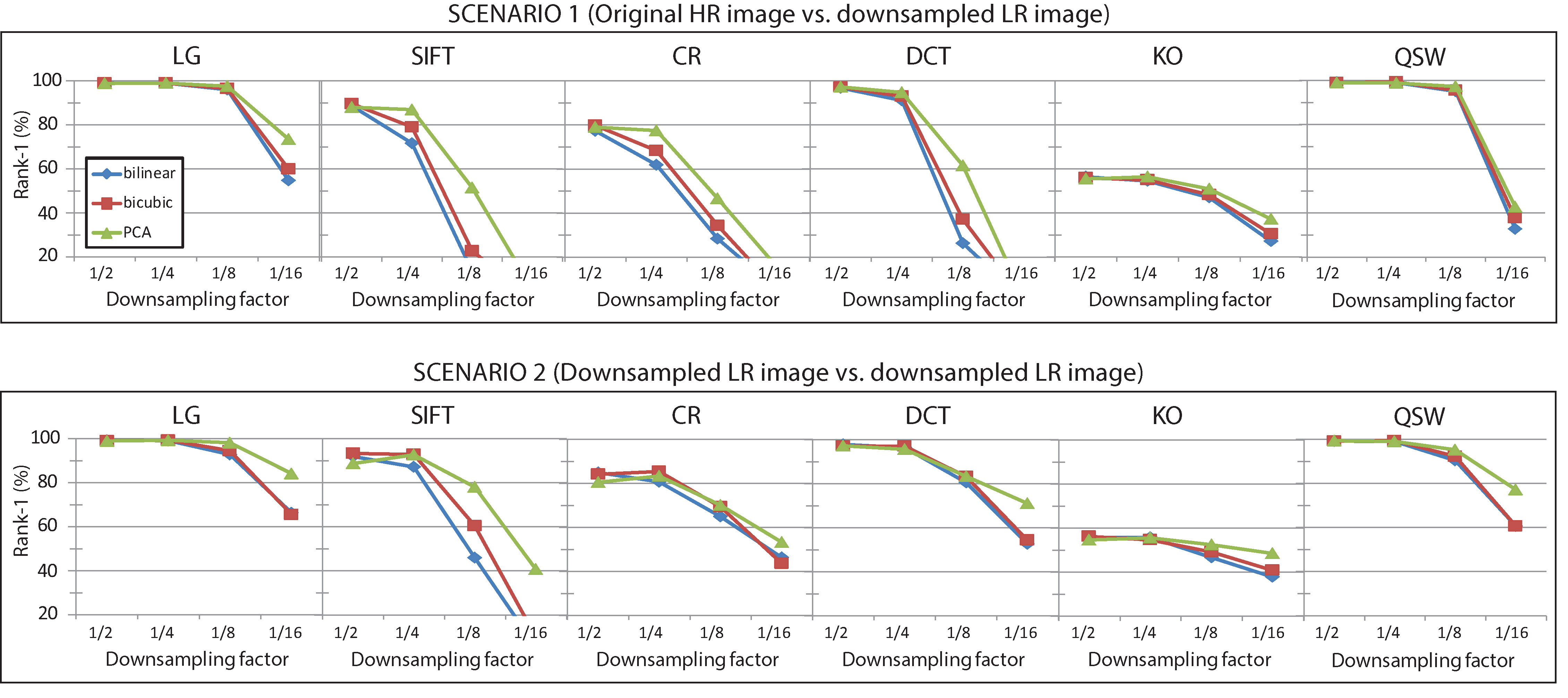}
     \caption{Identification results (Top-1 accuracy) with different downsampling factors.}
     \label{fig:results-rank1}
\end{figure*}

\begin{figure*}[htb]
     \centering
     \includegraphics[width=.98\textwidth]{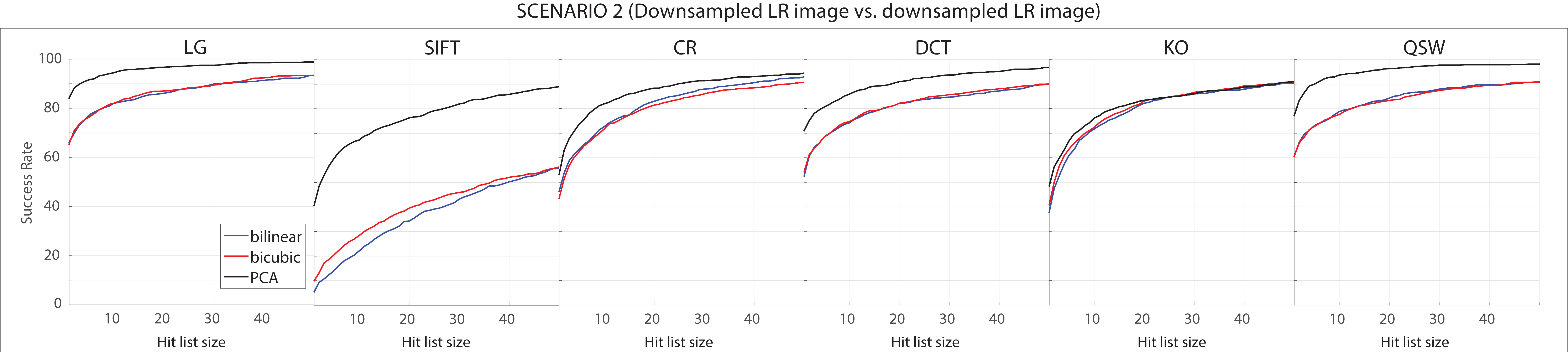}
     \caption{Identification results (CMC curves) of the iris comparators.
       Results are given for scenario 2 and a downsampling factor of 1/16 (image size 15$\times$15 pixels).}
     \label{fig:results-cmc}
\end{figure*}

\subsection{Biometric Identification}

In this section, we report identification experiments using hallucinated iris images.
%
Results
are given in Table~\ref{tabla:results-rank1} and Figure~\ref{fig:results-rank1}.
Here, we adopt as metric the classification accuracy for a hit list size of $k$=1 candidate
(also called Top-1 or Rank-1). 
In Table~\ref{tabla:results-rank1}, we also mark in bold the best accuracy for each comparator and for each downsampling factor.
In addition, the relative Top-1 accuracy variation of PCA with respect to bilinear/bicubic (best of the two) is given in brackets.
Finally, we provide in Figure~\ref{fig:results-cmc} the CMC curves
of scenario 2 for a downsampling factor of 1/16.

\subsubsection{Interpolation Method Analysis}

Similarly to the verification experiments, we also observe here a
superior performance of the trained PCA enhancement when the resolution decreases.
%
At an image size of 29$\times$29 pixels (downsampling of 1/8), the PCA method produces better accuracy than bilinear or bicubic interpolations for any comparator.
The Top-1 gain for this image size in scenario 1 reaches a value of 66.1\% (DCT comparator) and 124.5\% (SIFT), whereas in scenario 2, the gain reaches 29.3\% (SIFT).
In scenario 1, some comparators show a remarkable improvement even at a resolution of 57$\times$57 pixels. Here, CR has an improvement of 13.4\%, whereas SIFT has an improvement of 9.9\%.
But the advantage of using the PCA trained method becomes more evident again at very low resolution (15$\times$15 pixels), as it can be appreciated in the bigger separation between curves in Figure~\ref{fig:results-rank1}.
For example, in scenario 2, a Top-1 improvement of 309.1\% is observed with SIFT, followed by 30.8\% with DCT, and $\sim$27\% with LG and QSW.
In scenario 1, improvements of 483.7\% and 217.2\% can be observed with SIFT and CR, respectively, although the Top-1 accuracy of these comparators is far from being usable in practice (below 15\%).

\subsubsection{Scenario Analysis}

Regarding the two scenarios of operation employed,
similar observations than in the verification experiments can be made here.
The performance of a number of comparators (LG, KO, QSW, DCT) only differs between the two
scenarios when the downsampling factor is of 1/8 or higher, while the others (SIFT, CR) show differences
already at 1/4.
Also, when the two scenarios differ, the best performance is always obtained in scenario 2.
In scenario 1, only query samples are from low-resolution images, which are matched against high-resolution enrolment samples.
As we observed earlier, the similarity between a high-resolution image and its reconstructed version decreases with the resolution, which explains the worse performance obtained in scenario 1.
%
%

It is also specially relevant the poor performance of some comparators
in scenario 1 at a resolution of 1/16.
For example, SIFT and DCT obtain a Top-1 accuracy of $\sim$5\% or less. These same comparators also show a high EER in verification mode (above 36\%). This makes these two features unsuitable to compare images that have very different resolution, as it is the case in scenario 1. Indeed, the only comparator capable of handling such differences to some extent is LG. Its Top-1 identification accuracy in scenario 1 is of 73.43\%, and it also shows the best verification accuracy among all comparators (EER of 7.29\%).

\subsubsection{Comparator Analysis}

Finally, we analyse the differences between the individual comparators
when they operate in identification mode.
An effect observed in the verification experiments that we also see here is that
the performance of any comparator
remains without significant degradation until a downsampling factor of at least 1/8.
This corresponds to an image size of 29$\times$29 pixels.
Some comparators (like LG or QSW), are not significantly degraded even with this small image size,
having a Top-1 accuracy of 95-98\% in any of the two scenarios.
%

In terms of absolute performance, LG and QSW are, again, the best comparators, with a
Top-1 accuracy higher than 98\% across a great range of resolutions.
They only degrade at very low-resolution (15$\times$15 pixels),
showing a noteworthy Top-1 accuracy of 84.16\% (LG) and 77.23\% (QSW) in scenario 2.
%
The Top-1 accuracy of DCT at this very low-resolution
is 71.07\%, while accuracy of the other comparators fall below 53\%,
making them unfeasible for practical applications in this extreme case.

Regarding the CMC curves at very low-resolution (Figure~\ref{fig:results-cmc}),
if we allow a hit list size of $k$=5 candidates, both LG and QSW show a classification accuracy
higher than 90\%, which goes up to 95\% for a list of $k$=10 candidates.
This shows that the proposed PCA approach can be effectively used to improve
iris identification under severe downsampling.
With bilinear or bicubic interpolations,
the accuracy of these two comparators is below 78\% (if $k$=5)
and 82\% (if $k$=10)
Also, the DCT and CR comparators can reach 90\% accuracy with PCA enhancement,
but we need to increase the size of the hit list to
$k$=20 or $k$=25 candidates, respectively.
KO and SIFT, on the other hand, need a list size of $k$=50 candidates to reach 90\% accuracy.


\section{Conclusions}
\label{sect:conclusions}

Iris is regarded as one of the most accurate biometric modalities \cite{[Jain16]}.
It provides very high accuracy in controlled environments,
but deployment in non-controlled environments such as at-a-distance
or on-the-move is not yet mature \cite{[Nigam15]}.
The use of more relaxed acquisition environments is pushing image-based
biometrics towards the use of low-resolution imagery. This can pose significant
problems in terms of reduced performance if not tackled properly.
In this context,
super-resolution techniques
can be used to enhance the quality of low-resolution iris images and therefore, to improve the
recognition performance of existing systems \cite{[Nguyen18]}.

Super-resolution is a core topic in computer vision,
with many techniques proposed to restore
low-resolution images \cite{[Nasrollahi2014],[Thapa16]}.
However,
compared with the existing literature in generic super-resolution,
super-resolution in biometrics is a relatively recent topic \cite{[Nguyen18a]}.
%
%
%
This is because most approaches are general-scene, designed
to produce an overall visual enhancement.
They try to improve the quality of the image by minimizing an objective fidelity measure,
such as the Peak Signal-to-Noise Ratio (PSNR),
which does not necessarily correlate with better recognition performance \cite{[Nguyen12]}.
%
%
Images from a specific biometric modality have particular local and global structures that can be exploited to achieve a more efficient upsampling \cite{[Baker02]}.
For example, recovering local texture details is essential for iris images
due to the prevalence of texture-based recognition in this modality \cite{[Bowyer13handbookirissurvey]}.

In this paper, we present an extensive up-to-date survey of super-resolution applied to iris biometrics (Section~\ref{sect:soa}). We provide a comprehensive coverage of the existing literature, including a taxonomy of existing iris super-resolution approaches (Figure~\ref{fig:taxonomy} and Table~\ref{tabla:SOA}). They can be broadly classified into reconstruction-based and learning-based methods \cite{[Park03]}. Reconstruction based methods register and fuse a sequence of low-resolution images by pixel-wise combination of intensity values, in order to estimate a high-resolution image.
On the other hand, learning based methods use coupled dictionaries to learn the mapping relations between low- and high-resolution image pairs. 
The research community has lately focused on the latter category, since they can provide higher quality images and larger magnification factors \cite{[Farrugia17]}.
They also have the advantage of needing only one image as input.
In our survey, we also cover other aspects, from the database employed in the evaluation of each method, to the domain of operation (pixel or feature domain), the data used as input (iris images, polar images, or feature representations), the use of local patches, the smallest size of the input low-resolution image, or the biometric authentication experiments reported in each study.
The majority of works have employed near-infrared (NIR) data. 
Also, the majority of works reconstruct the polar representation of the iris image \cite{[Daugman04]}. 
It is also very common, specially among learning-based methods, to restore images at the patch level. 
Each patch is allowed to have its optimal reconstruction coefficients, helping to better recover local texture details, which are essential in iris recognition \cite{[Farrugia17]}.
%
%
Regarding the smallest size of the input low-resolution image, it is common to employ databases where subjects have been naturally captured at a certain distance, such as the MBGC portal, CASIA Mobile, or Q-FIRE databases.
Images in these databases have an average iris diameter in the range from 90 to 130 pixels.
This is in the limit of the minimum diameter recommended as sufficient for iris recognition, which has been found by experimental studies to be $\sim$120 pixels \cite{[Tabassi11]}.
Due to the lack of databases with smaller resolutions, some works carry out an artificial downsampling to achieve an smaller image size, which is a common approach in the super-resolution literature \cite{[Wang14]}.
In these studies, the average iris diameter employed is in the range from 11 to 53 pixels.

In the present paper,
we also investigate the use of a trained super-resolution enhancement technique
based on dictionary learning to
improve the resolution of near infrared iris images.
%
We study in depth a technique based on PCA Eigen-transformation of
local image patches (eigen-patches), inspired by the system
of \cite{[Chen14]} for face images.
Iris images are first resized and aligned such that the eye centre and the sclera are aligned.
Then, a dictionary is built for each patch position, which
is done by applying Principal Component Analysis to a set of collocated patches of low-resolution iris images from a training set.
During testing, given a low-resolution patch, it is first projected onto the low-dimensional eigen-space. Then, the reconstruction weights are used to restore the high-resolution patch using collocated patches contained in the coupled high-resolution training set.
In the employed method, the structure of the iris image is exploited in two ways: $i$) by building a patch position-dependent dictionary, which caters for a specific region of the iris; and $ii$) by allowing that each patch has its own optimum reconstruction weights, so the solution is locally optimized.
In addition,
unlike other methods that reconstruct images in the polar domain, we reconstruct the iris image directly, although our method is general enough to be applicable to polar images too. This makes our approach agnostic of the feature extraction method employed, given than there are comparators which do not transform the iris image to the polar domain, e.g. \cite{[Alonso09],[Nguyen18]}.

To evaluate the proposed method,
we conducted extensive 
experiments
with a database of 1,872 iris images.
Low-resolution images were simulated by downsampling high-resolution
irises. 
%
High-resolution images, of size 231$\times$231, were sub-sampled by $1/n$, with $n \in \left\{ {2,4,6,8,10,12,14,16} \right\}$.
%
This resulted in low-resolution images of
115$\times$115, 57$\times$57, 39$\times$39, 29$\times$29, 23$\times$23,
19$\times$19, 17$\times$17, and 15$\times$15 pixels respectively.
The latter corresponds to an iris diameter of 13 pixels.
%
Such a small resolution has not been previously employed in any iris super-resolution study, apart from ours.
This paper expands our previous studies \cite{[Alonso15b],[Alonso16b]} with a more comprehensive experimental framework. For the authentication experiments, we used publicly available feature extraction methods from popular and state-of-the-art schemes \cite{[Rathgeb13]}.
In particular, we included in this article four new iris recognition algorithms \cite{[Rathgeb10],[Monro07],[Ko07],[Ma03]},
which have been added to the two employed in our previous work \cite{[MasekThesis03],[Lowe04]}.
We also report identification experiments,
which are lacking in the majority of iris super-resolution studies.

Our experimental section starts by reporting the Peak
Signal-to-Noise Ratio (PSNR) and the Structural Similarity index (SSIM) between
the hallucinated and the corresponding high-resolution reference images.
The proposed method was compared with bilinear and bicubic interpolation.
%
%
Our experiments show the superiority of the presented PCA approach over these
two interpolation methods.
%
As resolution decreases, the gain of the PCA method over
bilinear or bicubic becomes more prominent, at least as measured by the PSNR.
At a resolution of 115$\times$115 pixels, the PSNR of PCA is, on average, 0.61dB higher than the PSNR of bicubic over the whole image. At 15$\times$15 pixels, the difference increases to 1.46dB.
This is consistent with a subjective assessment of the reconstructed images (Figure~\ref{fig:images-example}). When the resolution decreases, we observe that images reconstructed with bilinear or bicubic methods have much more blur than those reconstructed
with PCA.
On the other hand, this subjective difference is not captured by the SSIM.
As we observe in our experiments, the SSIM gain of PCA over bilinear or bicubic remains around 0.01-0.02 across the whole range of resolutions evaluated.
One of the drawbacks of these image fidelity metrics is precisely that they are not expected to have the same sensitivity to image degradations \cite{[AlNajjar12],[Hore10]}.
In our case, the SSIM is not a good predictor of the subjective differences observed between images reconstructed with the PCA algorithm and with the bilinear/bicubic methods.
Another drawback acknowledged in the literature is that
while they may reflect the \emph{goodness}
of the enhancement (in the visual sense), they are not necessarily good
predictors of the recognition accuracy \cite{[Nguyen12]}.
In other words, an overall visual enhancement of the image does not necessarily correlate with a better performance when such image is used for recognition purposes \cite{[Alonso12a]}.

Consequently,
to evaluate the usefulness of the proposed PCA reconstruction method to improve authentication accuracy,
we conducted a series of verification and identification experiments
with six different iris recognition algorithms based on different features.
We considered
two operational scenarios: one where original high-resolution
images are matched against hallucinated images
(`controlled' enrolment scenario, or scenario 1),
and another scenario where only
hallucinated images are used (`uncontrolled' scenario, or scenario 2).
The benefit of our trained approach becomes evident in comparison to
bilinear or bicubic interpolations when the resolution becomes very low.
%
For example, with an image size of 29$\times$29 pixels or smaller,
the PCA method gives better accuracy than bilinear or bicubic interpolations for any given comparator, highlighting the superiority of our trained PCA approach to enhance very low-resolution iris images.
With some comparators, the superiority of PCA is also appreciable even at a resolution of 57$\times$57 pixels.
%
%
%
%
It is also worth noting the resilience of some comparators to
severe downsampling (image size of 15$\times$15 pixels) when using PCA reconstruction.
Two particular comparators based on Gabor wavelets showed impressive EER values of
$\sim$5\% and a Top-1 accuracy of 77-84\% for this extreme case.

Another observation is
the different behaviour exhibited by the comparators employed when resolution is reduced.
The two mentioned Gabor-based comparators do not show a significant degradation in performance until an image size of 15$\times$15 pixels is used.
The majority of comparators, on the other hand, show appreciable degradation at a resolution of 29$\times$29 pixels. However, one particular comparator shows a high resilience to changes in resolution, with an EER in the range of 12-13\% and a Top-1 accuracy of 48-55\% for any given resolution.
This reinforces our previous assumption that enhancing an image in the visual sense,
or measured by a metric of \emph{fidelity},
does not necessarily contribute to a better recognition performance.
Since recognition algorithms are based on different features, and their
performance is not affected by the same factors, a metric of image fidelity
may be useful for a particular algorithm only \cite{[Alonso12a]}.
For this reason, focusing on the recognition performance of the algorithms employed becomes necessary.

Another interesting phenomenon is that since the performance of most comparators
is not significantly affected until an image size of 29$\times$29 pixels,
it would be feasible to use both query and test images of reduced size.
They could be then up-sampled simply with bicubic or bilinear interpolation prior to recognition.
This is of importance for example
in scenarios where storage or data transmission capacity is limited.
We also observe that the performance in scenario 2 is better than in scenario 1 for any comparator,
specially at low resolutions.
To further study this phenomenon, we looked into the score distributions of each comparator,
observing that genuine scores shift towards the impostor ones in scenario 1 as resolution decreases, a effect not observed in scenario 2.
In scenario 1, enrolment samples were taken from high-resolution images, whereas query samples were taken from reconstructed images. As resolution decreases, it is understandable that the similarity between the two types of images decreases as well, which explains the shift of the distributions of genuine scores.
%
%
In scenario 2, both enrolment and query images are reconstructed images, therefore it is expected that their relative similarity does not change to the same extent than in scenario 1.
%
%

The proposed enhancement algorithm assumes that
hallucination weights are the same in the low- and
high-resolution manifolds.
While this simplifies the problem, the low-resolution manifold
is usually distorted by the one-to-many
relationship between low- and high-resolution patches,
so this assumption may not hold always true \cite{[Wang14]}.
Another simplification is the assumption of
linearity in the combination of patches from the training dictionary.
In future work, we will explore methods to remove these two simplifications
in order to increase performance of
the hallucination algorithm.
In our most recent work, for example, we are adapting methods
which simultaneously consider the low- and high-resolution
manifolds during the hallucination process \cite{[Alonso17]}.
We will also look into strategies to iteratively
update the low-resolution dictionary
to reduce the modality gap between low- and high-resolution
patches contained within the dictionary \cite{[Jiang14]}.
Augmentation of the training set is another strategy that
we are considering by adding spatial offset during
the extraction of collocated patches, allowing to better cope
with eye alignment inaccuracies
and local image distortions.
Lastly, we are also studying the feasibility of
deep learning methods to provide an end-to-end mapping between low- and
high-resolution iris images \cite{[Ribeiro17],[Ribeiro18]}.

\bibliographystyle{IEEEbib}

\clearpage
\newpage

\begin{IEEEbiography}[{\includegraphics[width=1in,height=1.25in,clip,keepaspectratio]{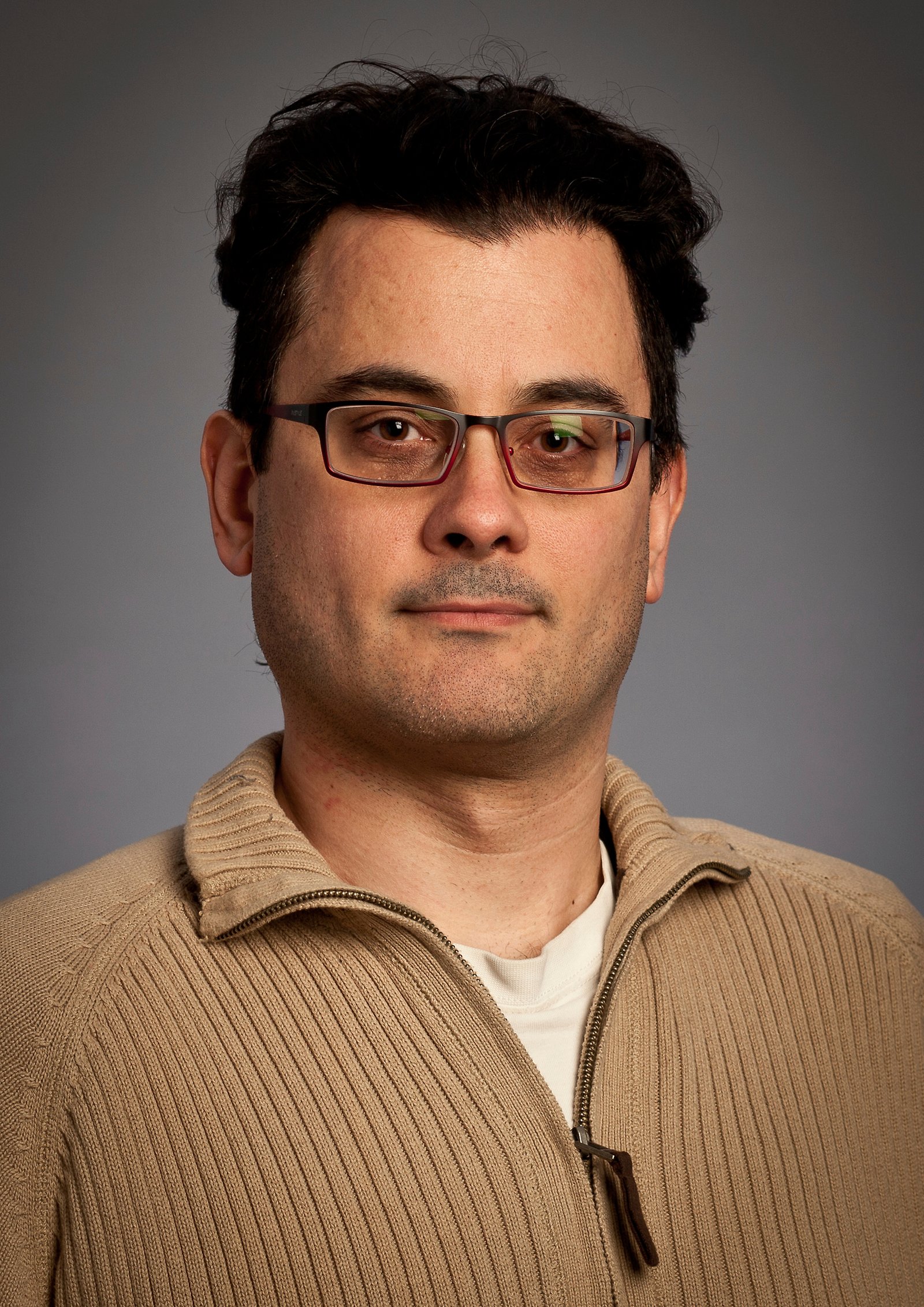}}]{Fernando Alonso-Fernandez}
received the M.S. and Ph.D. degrees in telecommunications engineering
from the Universidad Politecnica de Madrid, Spain, in 2003 and 2008, respectively.
Since 2010, he has been with the Centre for Applied Intelligent Systems Research,
Halmstad University, Sweden, first as the recipient of a Marie Curie IEF and a postdoctoral
fellowship from the Swedish Research Council, and later as the recipient of a Project Research Grant for
Junior Researchers of the Swedish Research Council. Since 2017, he is an Associate Professor at
Halmstad University.
He has been actively involved in in multiple EU (e.g. FP6 Biosecure NoE, COST IC1106)
and National projects focused on biometrics and human-machine interaction.
He also co-chaired ICB2016, the 9th IAPR International Conference on Biometrics.
His research interests include signal and image processing, pattern recognition, and biometrics, with emphasis on
facial cues and body biosignals. He has over 70 international contributions at refereed conferences and journals and has authored several book chapters.
Since 2018, he is in the Editorial Board of IET Biometrics and an Associate Editor of the IEEE Biometrics Council newsletter.
\end{IEEEbiography}

\begin{IEEEbiography}[{\includegraphics[width=1in,height=1.25in,clip,keepaspectratio]{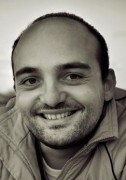}}]{Reuben A. Farrugia} (S'04, M'09, SM'17)
received the first
degree in electrical engineering in 2004, and the Ph.D.
degree in 2009, both from University of Malta, Msida,
Malta.
In January 2008, he was appointed Assistant
Lecturer with the same department and is currently
a Senior Lecturer. He has been in technical and organizational
committees of several national and international
conferences. In particular, he served as
the General-Chair on the IEEE International Workshop
on Biometrics and Forensics and as Technical
Programme Co-Chair on the IEEE Visual Communications and Image Processing
in 2014.
He has been contributing as a
Reviewer of several journals and conferences, including the IEEE Transactions
on Image Processing, IEEE Transactions on Circuits and
Systems for Video and Technology, and IEEE Transactions on Multimedia.
On September 2013, he was appointed as the National Contact Point of
the European Association of Biometrics.
\end{IEEEbiography}

\begin{IEEEbiography}[{\includegraphics[width=1in,height=1.25in,clip,keepaspectratio]{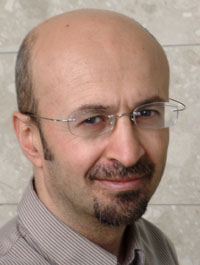}}]{Josef Bigun}
(M'88, SM'98) received the M.S. and
Ph.D. degrees from Linköping University, Sweden,
in 1983 and 1988, respectively.
Between 1988 and 1998, he was with the EPFL,
Switzerland. He was elected Professor to the Signal
Analysis Chair at Halmstad University and the
Chalmers University of Technology in 1998. His
scientific interests include a broad field in computer
vision, texture and motion analysis, biometrics,
and the understanding of biological recognition
mechanisms.
Dr. Bigun has co-chaired several international conferences. He has been contributing
as a referee or as an editorial board member of journals including Pattern
Recognition Letters and the IEEE Transactions
on Image Processing.
He served on the executive committees of several associations, including IAPR.
He has been elected Fellow of IAPR.
\end{IEEEbiography}

\begin{IEEEbiography}[{\includegraphics[width=1in,height=1.25in,clip,keepaspectratio]{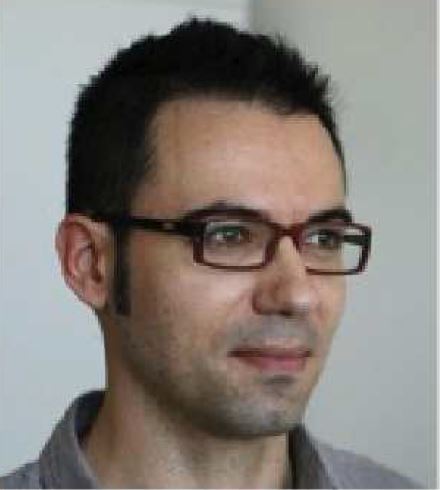}}]{Julian Fierrez} received the MSc and the PhD degrees in electrical engineering from Universidad Politecnica de Madrid, Spain, in 2001 and 2006, respectively. Since 2002 he has been affiliated with the Biometric Recognition Group at Universidad Autonoma de Madrid, where he is an Associate Professor since 2010. From 2007 to 2009 he was a postdoc at Michigan State University under a Marie Curie fellowship.
Prof. Fierrez has been actively involved in the last 15 years in large EU projects focused on biometrics (e.g. BIOSECURE, TABULA RASA and BEAT), and is the recipient of a number of distinctions, including the EAB European Biometric Industry Award 2006, the EURASIP Best PhD Award 2012, and the IAPR Young Biometrics Investigator Award 2017.
Since 2016 he is an Associate Editor for IEEE Trans. on Information Forensics and Security and the IEEE Biometrics Council newsletter.
\end{IEEEbiography}

\begin{IEEEbiography}[{\includegraphics[width=1in,height=1.25in,clip,keepaspectratio]{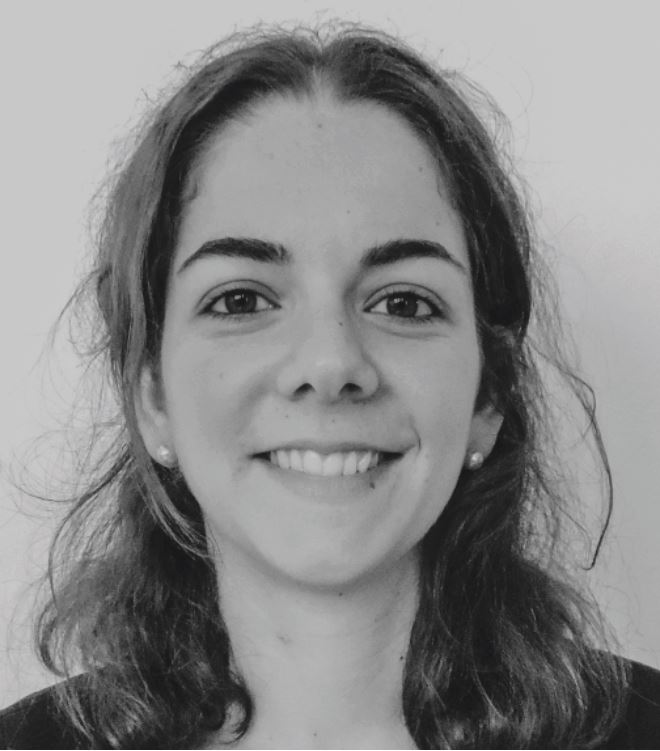}}]{Ester Gonzalez-Sosa} received the B.S. in computer
science and M.Sc in electrical engineering
from Universidad de Las Palmas de Gran
Canaria in 2012 and 2014, respectively. In June
2017 she obtained her PhD degree from Universidad
Autonoma de Madrid, within the Biometric
Recognition Group.
She has carried out several
research internships in worldwide leading
groups in biometric recognition such as TNO,
EURECOM, or Rutgers University. Her research
interests include pattern recognition, signal processing,
and biometrics, with emphasis on face, body, soft biometrics
and millimetre imaging. Gonzalez-Sosa has been the recipient of
UNITECO AWARD from the Spanish Association of Electrical Engineers,
the competitive Obra Social La CAIXA Scholarship, and the European Research Award 2018
of the European Association for Biometrics (EAB).
\end{IEEEbiography}

\EOD

\end{document}